\documentclass[sigconf]{acmart}
\copyrightyear{2026}
\acmYear{2026}
\setcopyright{cc}
\setcctype{by}
\acmDOI{XXXXXXX.XXXXXXX}
\acmConference[KDD 2026] {Proceedings of the 32nd ACM SIGKDD Conference on Knowledge Discovery and Data Mining V.2}{August 9--13, 2026}{Jeju Island, Republic of Korea.}
\acmBooktitle{Proceedings of the 32nd ACM SIGKDD Conference on Knowledge Discovery and Data Mining V.2 (KDD 2026), August 9--13, 2026, Jeju Island, Republic of Korea}
\acmISBN{979-8-4007-2259-2/2026/08}
\acmDOI{10.1145/3770855.3817466}
\usepackage[utf8]{inputenc}
\usepackage{hyperref}
\usepackage[most]{tcolorbox}
\usepackage{xspace}
\usepackage{enumitem}
\usepackage{subcaption} 
\usepackage{wrapfig}
\usepackage{xcolor}
\usepackage{colortbl}
\usepackage{makecell}
\usepackage{appendix}
\usepackage{caption}
\usepackage{multirow}
\usepackage{colortbl}
\usepackage{graphicx}
\usepackage[ruled,linesnumbered]{algorithm2e}
\usepackage{multicol}
\usepackage{booktabs}

\usepackage{amssymb}
\usepackage{pifont}

\definecolor{mygray}{gray}{.88}
\definecolor{mycyan}{cmyk}{.15,0,0,0}
\definecolor{mycyan2}{cmyk}{.85,0,0,0}
\definecolor{mygreen}{rgb}{0.19, 0.79, 0.02}
\definecolor{midnightgreen}{rgb}{0.0, 0.29, 0.33}
\definecolor{darkgreen}{RGB}{0,160,0}
\definecolor{backyellow_soft}{rgb}{1.0,1.0,0.8}
\definecolor{backred}{RGB}{255, 190, 190}
\definecolor{backblue}{RGB}{210, 230, 250}

\newcommand{\bench}{\textsc{LocalSearchBench}\xspace}
\newcommand{\rag}{\textsc{LocalRAG}\xspace}
\newcommand{\eval}{\textsc{LocalPlayground}\xspace}

\newcommand{\notcheckmark}{\textcolor{black}{\bcmark\kern-1.1ex\raisebox{.7ex}{\rotatebox[origin=c]{125}{--}}}\color{black}}
\newcommand{\bcmark}{\color{blue}{\ding{51}}}
\newcommand{\cmark}{\color{darkgreen}{\ding{51}}}
\newcommand{\xmark}{\color{red}{\ding{55}}}

\definecolor{todocolor}{rgb}{0.9,0.1,0.1}

\newcommand{\eg}{\hbox{\emph{e.g.}}\xspace}
\newcommand{\ie}{\hbox{\emph{i.e.}}\xspace}

\usepackage{pythonhighlight}

\definecolor{codegreen}{rgb}{0,0.6,0}
\definecolor{codegray}{rgb}{0.5,0.5,0.5}
\definecolor{codepurple}{rgb}{0.58,0,0.82}
\definecolor{backcolour}{rgb}{0.97,0.97,0.95}
\definecolor{forestgreen}{rgb}{0.28,0.62,0.37}

\lstdefinestyle{mystyle}{
    backgroundcolor=\color{backcolour},   
    commentstyle=\color{codegray},
    keywordstyle=\color{codepurple},
    numberstyle=\tiny\color{codegray},
    stringstyle=\color{blue},
    basicstyle=\ttfamily\footnotesize,
    breakatwhitespace=false,         
    breaklines=true,                 
    captionpos=b,                    
    keepspaces=true,                 
    numbers=left,                    
    numbersep=5pt,                  
    showspaces=false,                
    showstringspaces=false,
    showtabs=false,                  
    tabsize=4,
}

\begin{document}

\title{LocalSearchBench: Benchmarking Agentic Search in Real-World Local Life Services}
\settopmatter{authorsperrow=3}

\author{Hang He}
\affiliation{
  \institution{Meituan, Beijing, China}
  \city{}%
  \country{\kern0pt}%
  }
\affiliation{
  \institution{East China Normal University}
  \institution{Shanghai Innovation Institute}
  \city{Shanghai}
  \country{China}}
\email{hang.he@stu.ecnu.edu.cn}

\author{Chuhuai Yue}
\affiliation{ 
  \institution{Meituan, Beijing, China}
    \city{}%
  \country{\kern0pt}%
  }
\affiliation{
  \institution{Beijing Institute of Technology}
  \city{Beijing}
  \country{China}}
\email{chuhuaiyue@bit.edu.cn}

\author{Chengqi Dong}
\affiliation{
    \institution{Meituan, Beijing, China}
  \city{}%
  \country{\kern0pt}%
  }
\affiliation{
  \institution{University of Science and Technology of China}
  \city{Hefei}
  \country{China}}
\email{dongcq@mail.ustc.edu.cn}

\author{Mingxue Tian}
\affiliation{
    \institution{Meituan, Beijing, China}
  \city{}%
  \country{\kern0pt}%
  }
\affiliation{
  \institution{Shanghai Jiaotong University}
  \city{Shanghai}
  \country{China}}
\email{mingxuetian@sjtu.edu.cn}

\author{Hao Chen}
\affiliation{
    \institution{Meituan, Beijing, China}
  \city{}%
  \country{\kern0pt}%
  }
\affiliation{
  \institution{North China University of Technology, Beijing, China}
  \city{}%
  \country{\kern0pt}%
  }
\email{buycar@mail.ncut.edu.cn}

\author{Zhenfeng Liu}
\affiliation{
    \institution{Meituan, Beijing, China}
  \city{}%
  \country{\kern0pt}%
  }
\email{liuzhenfeng04@meituan.com}

\author{Jiajun Chai}
\authornote{Project leader.}
\affiliation{
    \institution{Meituan, Beijing, China}
  \city{}%
  \country{\kern0pt}%
  }
\email{chaijiajun@meituan.com}

\author{Xiaohan Wang}
\affiliation{
    \institution{Meituan, Beijing, China}
  \city{}%
  \country{\kern0pt}%
  }
\email{wangxiaohan17@meituan.com}

\author{Yufei Zhang}
\affiliation{
    \institution{Meituan, Beijing, China}
  \city{}%
  \country{\kern0pt}%
  }
\email{zhangyufei08@meituan.com}

\author{Qun Liao}
\affiliation{
    \institution{Meituan, Beijing, China}
  \city{}%
  \country{\kern0pt}%
  }
\email{liaoqun@meituan.com}

\author{Guojun Yin}
\authornote{Corresponding authors.}
\affiliation{
    \institution{Meituan, Beijing, China}
  \city{}%
  \country{\kern0pt}%
  }
\email{yinguojun02@meituan.com}

\author{Wei Lin}
\affiliation{
    \institution{Meituan, Beijing, China}   
    \city{}%
    \country{\kern0pt}%
    }
\email{linwei31@meituan.com}

\author{Chengcheng Wan}
\authornotemark[2]
\affiliation{
  \institution{East China Normal University}
  \institution{Shanghai Innovation Institute}
  \city{Shanghai}
  \country{China}}
\email{ccwan@sei.ecnu.edu.cn}

\author{Haiying Sun}
\affiliation{
  \institution{East China Normal University}
  \city{Shanghai}
  \country{China}}
\email{hysun@sei.ecnu.edu.cn}

\author{Ting Su}
\authornotemark[2]
\affiliation{
  \institution{East China Normal University}
  \city{Shanghai}
  \country{China}}
\email{tsu@sei.ecnu.edu.cn}

\renewcommand{\shortauthors}{Hang He et al.}


\begin{abstract}
Recent advances in large reasoning models (\emph{LRM}s) have enabled \emph{agentic search} systems to perform complex multi-step reasoning across multiple sources. However, most studies focus on general information retrieval and rarely explore vertical domains with unique challenges. In this work, we focus on local life services and introduce \bench, which encompasses diverse and complex business scenarios. Real-world queries in this domain are often ambiguous and require multi-hop reasoning across merchants and products, remaining challenging and not fully addressed. As the first comprehensive benchmark for \emph{agentic search} in local life services, \bench comprises a database of over 1.3M merchant entries across 6 service categories and 9 major cities, and 900 multi-hop QA tasks from real user queries that require multi-step reasoning. We also developed \eval, a unified environment integrating multiple tools for \emph{LRM}s interaction. Experiments show that even state-of-the-art \emph{LRM}s struggle on \bench: the best model (DeepSeek-V3.2) achieves only 35.60\% correctness, and most models have issues with completeness (average 60.32\%) and faithfulness (average 30.72\%). This highlights the need for specialized benchmarks and domain-specific agent training in local life services. 

\end{abstract}


\maketitle

\newcommand\kddavailabilityurl{https://doi.org/10.5281/zenodo.20416849}
\ifdefempty{\kddavailabilityurl}{}{
\begingroup\small\noindent\raggedright\textbf{KDD Availability Link:}\\
The source code of this paper has been made publicly available at \url{\kddavailabilityurl}.
\endgroup
}

\section{Introduction}
\sloppy

\begin{figure}[t]
    \centering
    \begin{subfigure}[b]{\columnwidth}
        \centering
        \includegraphics[width=\textwidth]{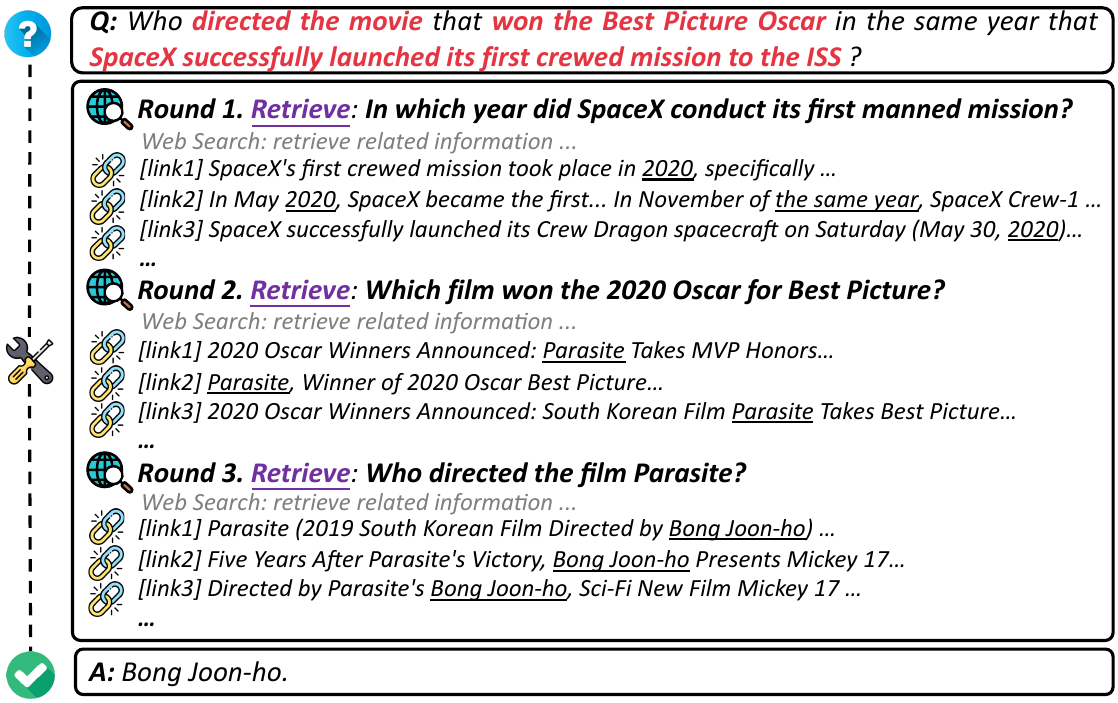}
        \caption{General multi-hop QA}
        \label{fig:intro_1}
    \end{subfigure}
    
    \begin{subfigure}[b]{\columnwidth}
        \centering
        \includegraphics[width=\textwidth]{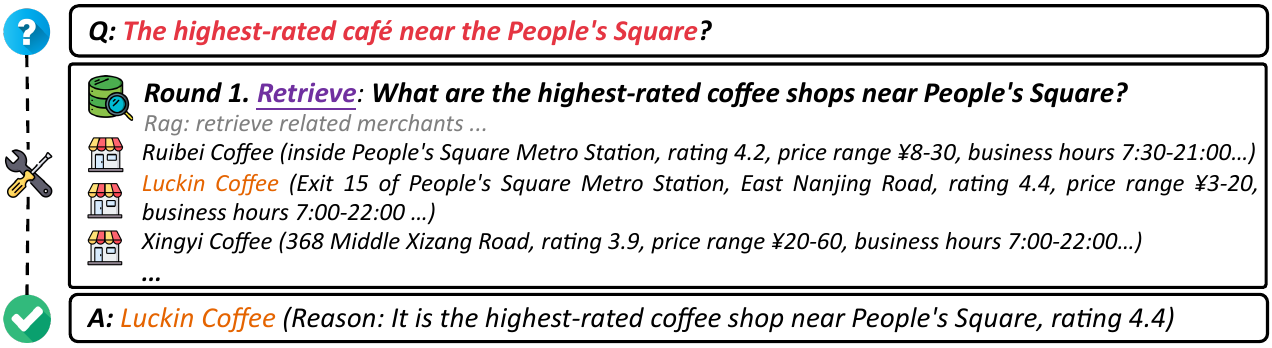}
        \caption{Local life service single-hop QA}
        \label{fig:intro_2}
    \end{subfigure}
    
    \begin{subfigure}[b]{\columnwidth}
        \centering
        \includegraphics[width=\textwidth]{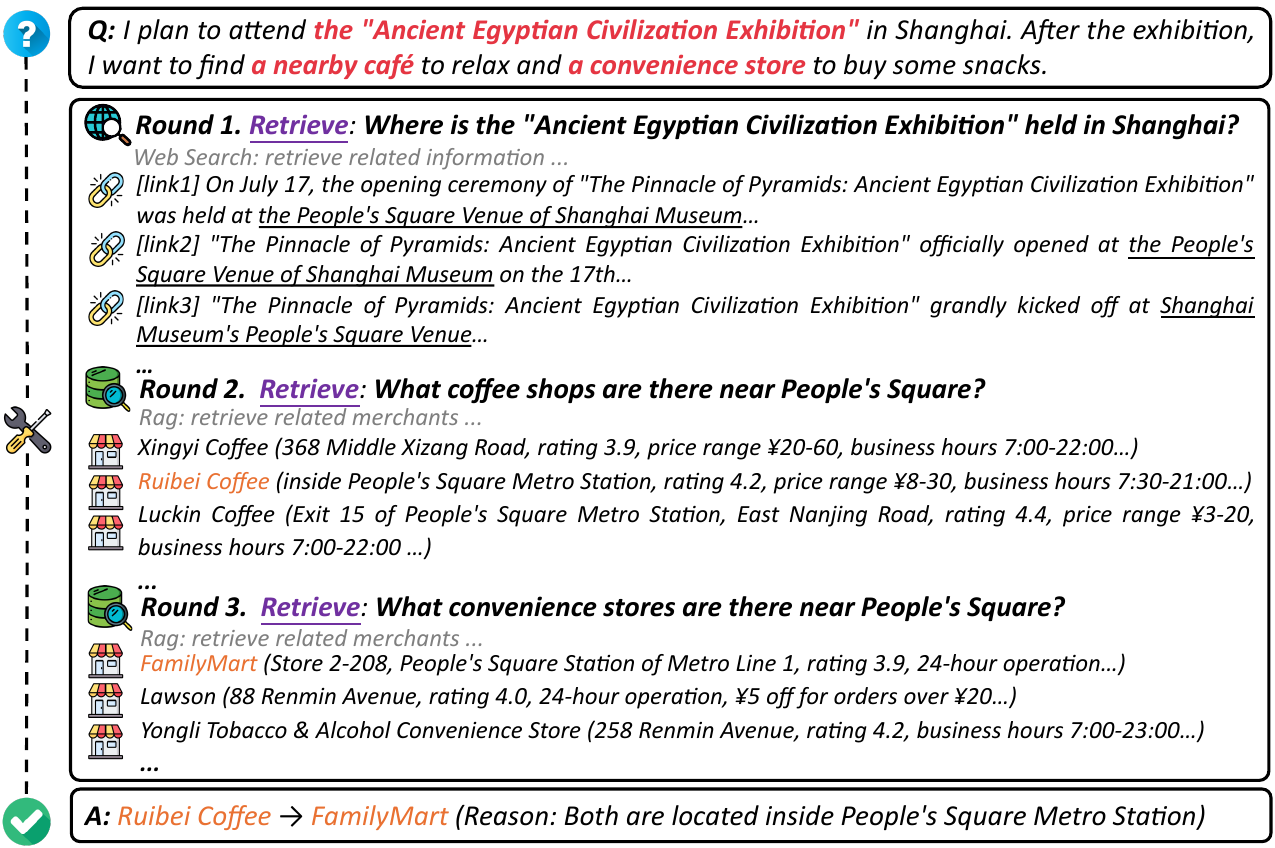}
        \caption{Local life service multi-hop QA}
        \label{fig:intro_3}
    \end{subfigure}
    \caption{Illustration of different types of QA tasks}
    \label{fig:intro}
    \vspace{-10pt}
\end{figure}

Recent advances in large reasoning models (\emph{LRM}s) have significantly enhanced the capabilities of AI agents, enabling them to perform complex reasoning and planning tasks~\cite{xu2025towards,ferrag2025llm,wu2025agentic,li2025webthinker,ke2025survey,chai2025rlfactory,chen2025toolforgedatasynthesispipeline}. This progress has driven the development of \emph{agentic search} systems, which can decompose complex queries, execute multi-step reasoning, and dynamically orchestrate multiple information sources to achieve autonomous cross-domain reasoning~\cite{sun2025simpledeepsearch,zhang2025evolvesearch,shi2025pangudeepdiveradaptivesearch,shi2025toollearningwildempowering,zhang2025web,li2025search,mo2025conversational,yue2025promoting,yue2025uiorchestra}. These systems also demonstrate the potential to expand from general domains to vertical application scenarios such as e-commerce~\cite{zhou2024webarenarealistic,yao2023webshopscalable}. However, there has been little work applying \emph{agentic search} to the local life service domain, and the lack of suitable evaluation benchmarks further constrains new research in this area. 

\emph{\textbf{Local life services}} are online-offline integrated services connecting local merchants with nearby users, including dining, lifestyle, shopping, accommodation, travel, healthcare and more to meet residents' daily needs~\cite{lan2025benchmarking,10.1145/3580305.3599874}. Leveraging location-based digital platforms, these services form complex networks of businesses, user preferences, and location-dependent service offerings, creating a vast market~\cite{10.1145/3534678.3539122,kusk2025flexible,liu2025mrgrp}. As shown in Figure~1(\subref{fig:intro_3}), real-world user queries often involve multi-constraint merchant recommendation, spatiotemporal service chain planning, and event-driven service bundling, which require complex multi-step processes to deliver the final response. 

\citet{lan2025benchmarking} pioneered systematic evaluation in this domain, but their benchmark focuses primarily on basic retrieval scenarios, leaving complex compositional queries under-explored. Moreover, existing \emph{agentic search} benchmarks (Table~\ref{tab:benchmark_comparison}) focus on general domains and miss critical local-life challenges such as geographic constraints and multi-service integration. This coverage gap highlights the urgent need for a multi-hop benchmark tailored to local life scenarios, accurately reflecting the complexity of real-world search tasks and fully challenging existing \emph{agentic search} methods.  

To bridge these gaps, we first introduce \textbf{\bench}, a comprehensive benchmark comprising over 1.3M data entries across 6 primary service categories, spanning 9 major cities in China. These entries are transformed from raw data into a high-quality augmented and anonymized database. Based on these data, we propose a high-quality instruction synthesis method to design 900 multi-hop QA tasks across two intelligence levels. These multi-hop QA tasks are constructed through a multi-agent data generation process, requiring agents to perform complex reasoning across multiple information sources and reasoning steps.

Building upon \bench, we further develop \textbf{\eval}, an \emph{agentic search} framework to evaluate LLM agents for local life services. Given the absence of appropriate benchmarks and the critical need for comprehensive evaluation systems, our approach addresses these challenges through comprehensive dataset construction, systematic evaluation framework and targeted model evaluation. We build a comprehensive evaluation environment with \rag based on the merchant database and web search tools to simulate real-world local life service search scenarios where models need to retrieve and synthesize information from multiple sources, and conduct evaluations of multiple \emph{LRM}s.

\begin{figure}[t]
    \centering
    \includegraphics[width=\linewidth]{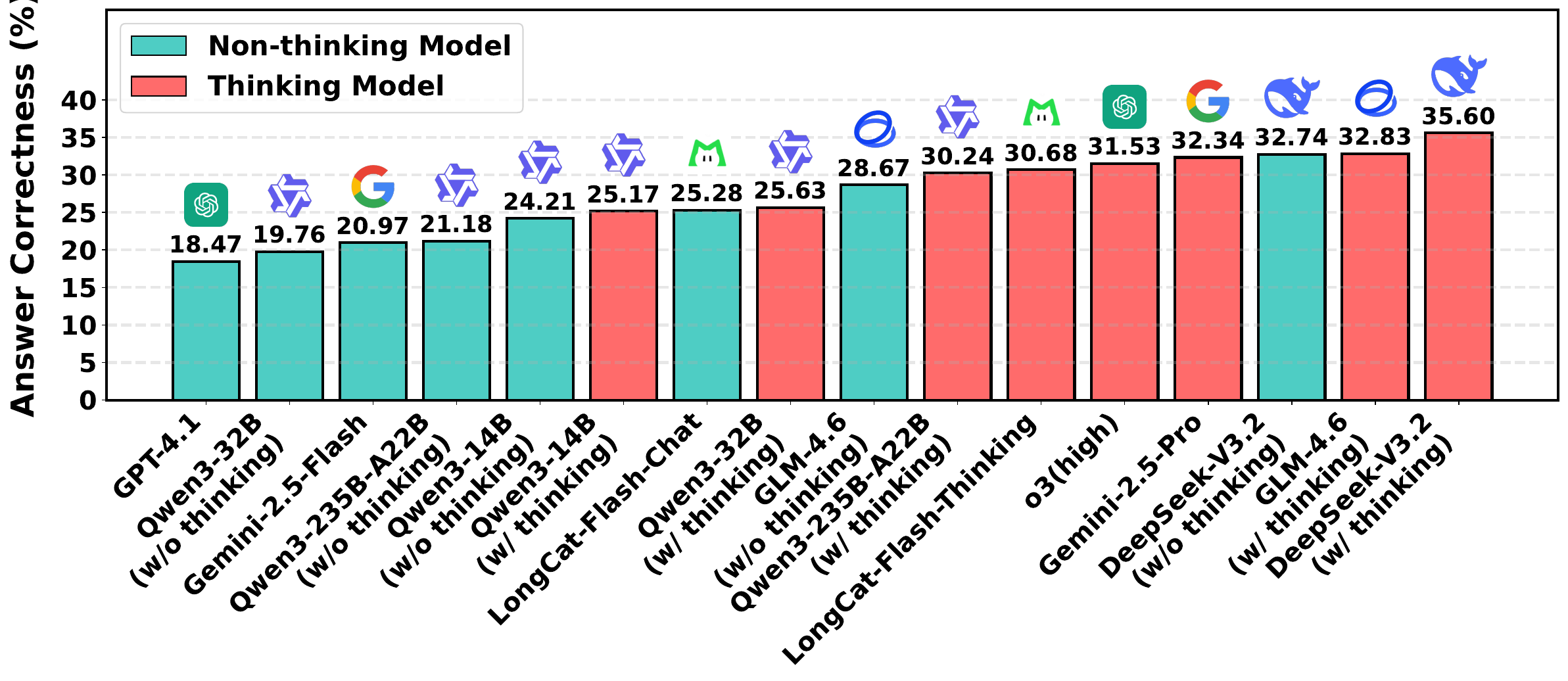}
    \caption{Performance comparison of various LLMs on answer correctness.}
    \label{fig:correctness_comparison}
    \vspace{-15pt}
\end{figure}

In summary, we make the following main contributions:
\begin{itemize}[leftmargin=*, itemsep=0pt, topsep=2pt, parsep=0pt]
 \item To the best of our knowledge, we are the first to build a comprehensive offline high-quality real-world agentic search benchmark for local life services.
 \item We propose a systematic \emph{agentic search} framework \eval and tool interaction environment for evaluating and applying \emph{LRM}s in local life services.
 \item We conduct extensive experiments across diverse models, revealing significant challenges in \textit{agentic search} on real-world tasks, and providing actionable insights for future research.
\end{itemize}

\section{Related Work}
\begin{table*}[h]
\centering
\caption{Comparison of existing user interaction benchmarks across key evaluation dimensions. ``{\cmark}''~indicates fully addressed, ``{\notcheckmark}''~indicates partially addressed, and ``{\xmark}''~indicates not addressed.}
\label{tab:benchmark_comparison}
\renewcommand{\arraystretch}{0.8}
\small
\resizebox{\linewidth}{!}{
\begin{tabular}{@{}lccccccccc@{}}
\toprule[1.5pt]
\multirow{2}{*}{\textbf{Benchmark}} & \multicolumn{3}{c}{\textbf{Reasoning Complexity}} & \multicolumn{2}{c}{\textbf{Domain Specificity}} & \multicolumn{3}{c}{\textbf{Task Complexity}} \\
\cline{2-9}
& \makecell{Multi-hop\\Reasoning} & \makecell{Tool\\Integration} & \makecell{Context\\Dependency} & \makecell{Domain\\Knowledge} & \makecell{Geographic\\Constraints} & \makecell{Multi-service\\Integration} & \makecell{Real-world\\Scenarios} & \makecell{Scale\\(samples)} \\
\midrule
MuSiQue~\cite{trivedi2022musique} & \cmark & \xmark & \cmark & \xmark & \xmark & \xmark & \xmark & 25K \\
BrowseComp~\cite{wei2025browsecomps} & \cmark & \cmark & \notcheckmark & \xmark & \xmark & \xmark & \notcheckmark & 1K \\
DeepResearch~\cite{du2025deepresearch} & \cmark & \cmark & \cmark & \notcheckmark & \xmark & \xmark & \cmark & 500 \\
DeepWideSearch~\cite{lan2025deepwidesearch} & \cmark & \cmark & \cmark & \notcheckmark & \xmark & \xmark & \cmark & 220 \\
L-MARS~\cite{wang2025lmars} & \cmark & \cmark & \cmark & \notcheckmark & \xmark & \xmark & \cmark & 200 \\
ProxyQA~\cite{tan2024proxyqa} & \cmark & \xmark & \cmark & \notcheckmark & \xmark & \xmark & \notcheckmark & 1550 \\
ResearchY~\cite{rosset2024researchy} & \cmark & \cmark & \cmark & \notcheckmark & \xmark & \xmark & \cmark & 1.5K \\
\midrule
\rowcolor{blue!10}
\textbf{\bench (ours)} & \cmark & \cmark & \cmark & \cmark & \cmark & \cmark & \cmark & \textbf{1.35M + 900} \\
\bottomrule[1.5pt]
\end{tabular}
}
\end{table*}

\subsection{Agentic Search} 
\citet{gou2025mind2web2evaluatingagentic} define \emph{agentic search} as systems capable of integrating multiple auxiliary tools to autonomously perform interactive generation through search tools (\eg, web search APIs or external databases) for handling complex search tasks. Leveraging the powerful language understanding capabilities of LLMs, these search agents can proactively address ambiguous or underspecified queries through clarification or reasonable assumptions~\cite{yang2025agentic}. They decompose the initial search task into manageable sub-queries, perform dynamic reasoning and planning based on accumulated contextual information, adjust search strategies in real time, and synthesize information from multiple sources~\cite{li2025towards}.

Recent advances in the reasoning abilities of \emph{LRM}s further foster the development of \emph{LRM}-based agentic search systems. The most representative products include OpenAI Deep Research~\cite{jaech2024openai}, Gemini Deep Research~\cite{comanici2025gemini}, and Tongyi Deep Research~\cite{tongyidr}. In addition to accessing real-time information through search APIs, these systems leverage various auxiliary tools to enable deeper and broader research on highly complex problems. In the open-source community, ~\citet{li2025search} first proposed integrating the agentic search workflow into the o1-style reasoning process of \emph{LRM}s to achieve autonomous knowledge supplementation. ~\citet{li2025webthinker}, ~\citet{jin2025search}, and ~\citet{song2025r1} further enhanced the autonomous search capabilities of LLMs by applying reinforcement learning, enabling the models to generate queries and invoke external retrieval systems during reasoning.
    
These methods have made significant progress in general information retrieval and question answering. However, there has been little advancement in vertical domains, especially in areas related to local lifestyle services, due to the challenges associated with training domain-specific agents and the lack of high-quality evaluation benchmarks. In this paper, we propose \bench, a comprehensive benchmark specifically designed for evaluating \emph{LRM}s in local life service scenarios. 

\noindent

\subsection{Benchmarking Agentic Capabilities}
Existing benchmarks for agentic search systems in the general domain can be broadly categorized into closed-ended and open-ended QA tasks. Closed-ended QA benchmarks include multi-hop reasoning datasets that require synthesizing information from multiple sources~\cite{yang2018hotpotqa,ho2020constructing,press2022measuring,trivedi2022musique}, challenging QA benchmarks designed for long-horizon questions with long-tail knowledge~\cite{wei2025browsecomps,zhou2025browsecompzhbenchmarking,xi2025infodeepseek,pham2025sealqa}, and fact-checking tasks that evaluate claim verification abilities~\cite{wei2024long,wadden2020fact,jiang2020hover,wang2024mfc}. Open-ended QA benchmarks focus on deep information seeking tasks, targeting multi-perspective queries~\cite{rosset2024researchy} and expert-level research tasks~\cite{du2025deepresearch,bosse2025deep,tan2024proxyqa}, with some incorporating multi-modal capabilities~\cite{yuan2025videodeepresearch,dong2025trainingmultiimagevisionagents,liu2025queries}. 

Despite covering various reasoning abilities, existing benchmarks differ from real user queries in local life services and lack assessment of multi-domain coordination. Users often face complex tasks requiring simultaneous service bookings and constraint satisfaction, which current datasets do not adequately evaluate.

Although these benchmarks cover a range of reasoning capabilities, they have a significantly different distribution from real user queries in local life service scenarios, lacking assessment of the multi-domain service coordination required in real-world applications. In reality, users often need to accomplish complex tasks that require coordinating multiple services, such as simultaneously booking restaurants, arranging transportation, and planning entertainment, all while satisfying geographic 
and temporal constraints. Existing specialized datasets are insufficient for evaluating challenging scenarios like event planning that require multiple API calls across service domains and real-time constraint satisfaction.

We propose \bench to assess multi-hop reasoning and coordinated retrieval for complex local life service tasks, and \eval as evaluation frameworks that comprehensively measure performance in these specialized scenarios.

\section{Construction of \bench}
\sloppy
\begin{figure*}[t] 
    \centering
    \includegraphics[width=\linewidth]{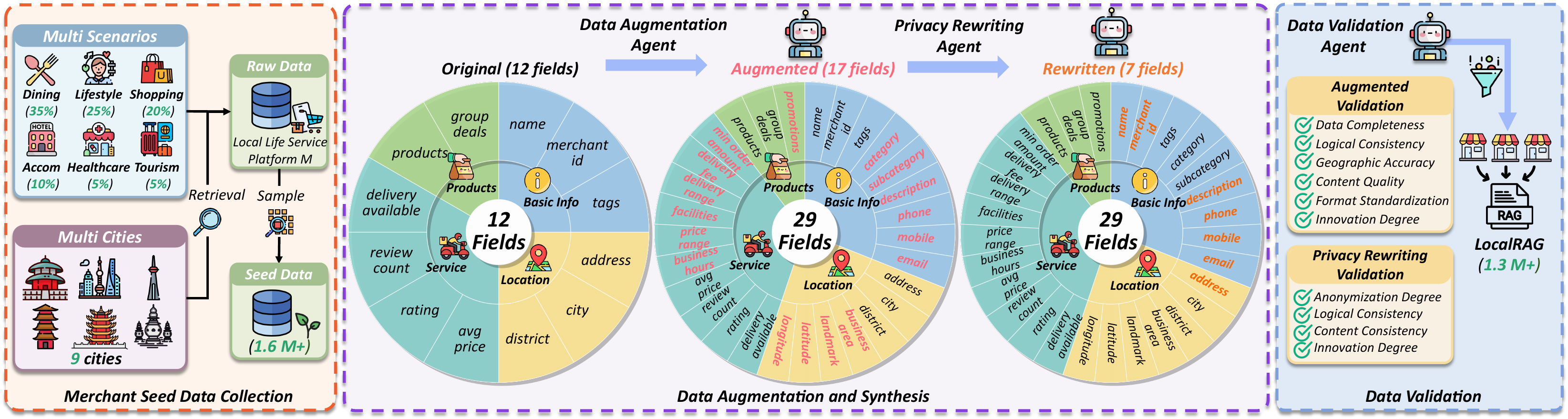}
    \caption{How Local Merchant Database in \bench are constructed.
    }
    \label{fig:seed}  
    \vspace{-5pt}
\end{figure*}

\bench is a benchmark designed for agentic search in local life services. It contains: (i) a local merchant database (\S~\ref{sec:database_construction}), (ii) a merchant search tool \rag (\S~\ref{sec:localrag}), and (iii) a multi-hop QA benchmark (\S~\ref{sec:multi_hop_qa}).

\subsection{Local Merchant Database Construction}
\label{sec:database_construction}
We build the local merchant database that underpins the multi-hop QA tasks in \bench. As shown in Figure~\ref{fig:seed}, the construction has three stages: \emph{Merchant Seed Data Collection}, \emph{Data Augmentation and Synthesis}, and \emph{Data Validation}.

\subsubsection{Merchant Seed Data Collection}
\label{lab:Seed Data Collection}
We build our merchant database from raw data of a leading local life service platform M~\cite{10.1145/3580305.3599874}, sampling over 1.6M merchant records from 2025 across 9 major Chinese cities. \label{subsubsection:Multi-Scenario}The collection covers six primary service categories: \textbf{Dining (35\%), Lifestyle (25\%), Shopping (20\%), Accommodation (10\%), Healthcare (5\%), and Tourism (5\%)} (Figure~9(\subref{fig:proportion})), with 927 keywords reflecting real search behaviors. \label{subsubsection:Multi-City Geographic}The geographic coverage spans 9 major cities with 10,741 landmarks covering commercial, transportation, cultural, educational, medical, and residential areas (Table~\ref{tab:geographic_coverage}). Figure~\ref{fig:city_ciyun} shows landmark distributions for 3 cities. Construction involved over 9.9M retrievals.

\begin{table}[htbp]
    \centering
    \caption{Geographic coverage across 9 major Chinese cities}
    \label{tab:geographic_coverage}
    \renewcommand{\arraystretch}{0.9}
    \resizebox{0.7\linewidth}{!}{%
    \begin{tabular}{c||c|c|c}
    \toprule[1.5pt]
    \rowcolor{gray!30}
    \textbf{City} & \textbf{Landmarks} & \textbf{Districts} & \textbf{Retrievals} \\
    \hline\hline
    Shanghai & 1556 & 16 & 1,442,412 \\
    \hline
    \rowcolor{gray!10}
    Beijing & 1451 & 16 & 1,345,077 \\
    \hline
    Guangzhou & 1026 & 11 & 951,102 \\
    \hline
    \rowcolor{gray!10}
    Shenzhen & 1039 & 9 & 963,153 \\
    \hline
    Hangzhou & 1129 & 13 & 1,046,583 \\
    \rowcolor{gray!10}
    \hline
    Chengdu & 1073 & 20 & 994,671 \\
    \hline
    Chongqing & 1008 & 38 & 934,416 \\
    \rowcolor{gray!10}
    \hline
    Wuhan & 1040 & 13 & 964,080 \\
    \hline
    Suzhou & 1149 & 9 & 1,065,123 \\
    \rowcolor{gray!30}
    \hline\hline
    \textbf{Total} & \textbf{10,741} & \textbf{145} & \textbf{9,956,907} \\
    \bottomrule[1.5pt]
    \end{tabular}
    \vspace{-25pt}
}
\end{table}

\subsubsection{Data Augmentation and Synthesis}
The original merchant seed data contains only 12 basic fields with many missing or unstandardized attributes, which is inadequate for agentic search tasks requiring dense merchant profiles. We employ a \emph{Data Augmentation Agent} to expand these records into 29 fields across 4 dimensions, enriching each merchant's basic, locational, service, and product information through 17 newly added fields. To ensure safe public release, we then employ a \emph{Privacy Rewriting Agent} to anonymize sensitive merchant data by rewriting 7 privacy-related fields (identity, location, contact details) while preserving business information, as illustrated in Figure~\ref{fig:seed}. Validation experiments show that data augmentation significantly improves retrieval quality (16.5\% gain in NDCG@10) while privacy rewriting causes minimal degradation (0.34\%--0.40\% across metrics), confirming that the entire pipeline preserves realistic search signals.

\subsubsection{Data Validation}
\label{sec:data_validation}
A \emph{Data Validation Agent} (LLM-as-judge) performs a two-stage quality assessment: (1) augmented data quality assessed on 6 weighted dimensions (Completeness, Logical Consistency, Geographic Accuracy, Content Quality, Format Standardization, Innovation Degree); (2) privacy-rewriting quality on 4 dimensions (Anonymization Degree, Logical Consistency, Content Consistency, Innovation Degree) 
(Table~\ref{tab:evaluation_comparison}).
\begin{table}[htbp]
\centering
\caption{Evaluation Frameworks for Data Quality Assessment}
\label{tab:evaluation_comparison}
\vspace{-10pt}
\begin{minipage}[htbp]{0.3\textwidth}
    \centering
    \subcaption{Augmented Data Quality}
    \label{tab:evaluation_result_1}
    \resizebox{\textwidth}{!}{%
    \begin{tabular}{c||c|c}
    \toprule[1.5pt]
    \rowcolor{gray!30}
    \textbf{Dimension} & \textbf{Weight} & \textbf{Score} \\
    \hline\hline
    Data Completeness & 20\% & 0.8605 \\
    \hline
    \rowcolor{gray!10}
    Logical Consistency & 30\% & 0.9000 \\
    \hline
    Geographic Accuracy & 10\% & 0.8568 \\
    \hline
    \rowcolor{gray!10}
    Content Quality & 20\% & 0.8521 \\
    \hline
    Format Standardization & 10\% & 0.9060 \\
    \hline
    \rowcolor{gray!10}
    Innovation Degree & 10\% & 0.7080 \\
    \hline\hline
    \rowcolor{blue!20}
    \textbf{Overall} & \textbf{100\%} & \textbf{0.8596} \\
    \bottomrule
    \end{tabular}%
    }
\end{minipage}
\hfill
\begin{minipage}[htbp]{0.3\textwidth}
    \centering
    \subcaption{Privacy Rewriting Data Quality}
    \label{tab:evaluation_result_2}
    \resizebox{\textwidth}{!}{%
    \renewcommand{\arraystretch}{0.9}
    \begin{tabular}{c||c|c}
    \toprule
    \rowcolor{gray!30}
    \textbf{Dimension} & \textbf{Weight} & \textbf{Score} \\
    \hline\hline
    Anonymization Degree & 50\% & 0.9473 \\
    \hline
    \rowcolor{gray!10}
    Logical Consistency & 20\% & 0.8383 \\
    \hline
    Content Consistency & 20\% & 0.9919 \\
    \hline
    \rowcolor{gray!10}
    Innovation Degree & 10\% & 0.8200 \\
    \hline\hline
    \rowcolor{blue!20}
    \textbf{Overall} & \textbf{100\%} & \textbf{0.9217} \\
    \bottomrule[1.5pt]
    \end{tabular}%
    }
\end{minipage}
\end{table}

\subsubsection{Local Merchant Database Statistics}
\begin{figure}
    \centering
    \setlength{\fboxsep}{0pt}
    \setlength{\fboxrule}{0.5pt}
    \resizebox{0.45\textwidth}{!}{%
    \begin{tabular}{@{}c@{\hspace{0.02\columnwidth}}c@{}}
        \begin{tabular}[c]{@{}c@{}}  
            \begin{subfigure}[b]{0.26\columnwidth}
                \centering
                \fbox{\includegraphics[width=\dimexpr\textwidth-2\fboxrule\relax, height=\dimexpr\textwidth-2\fboxrule\relax, keepaspectratio=false]{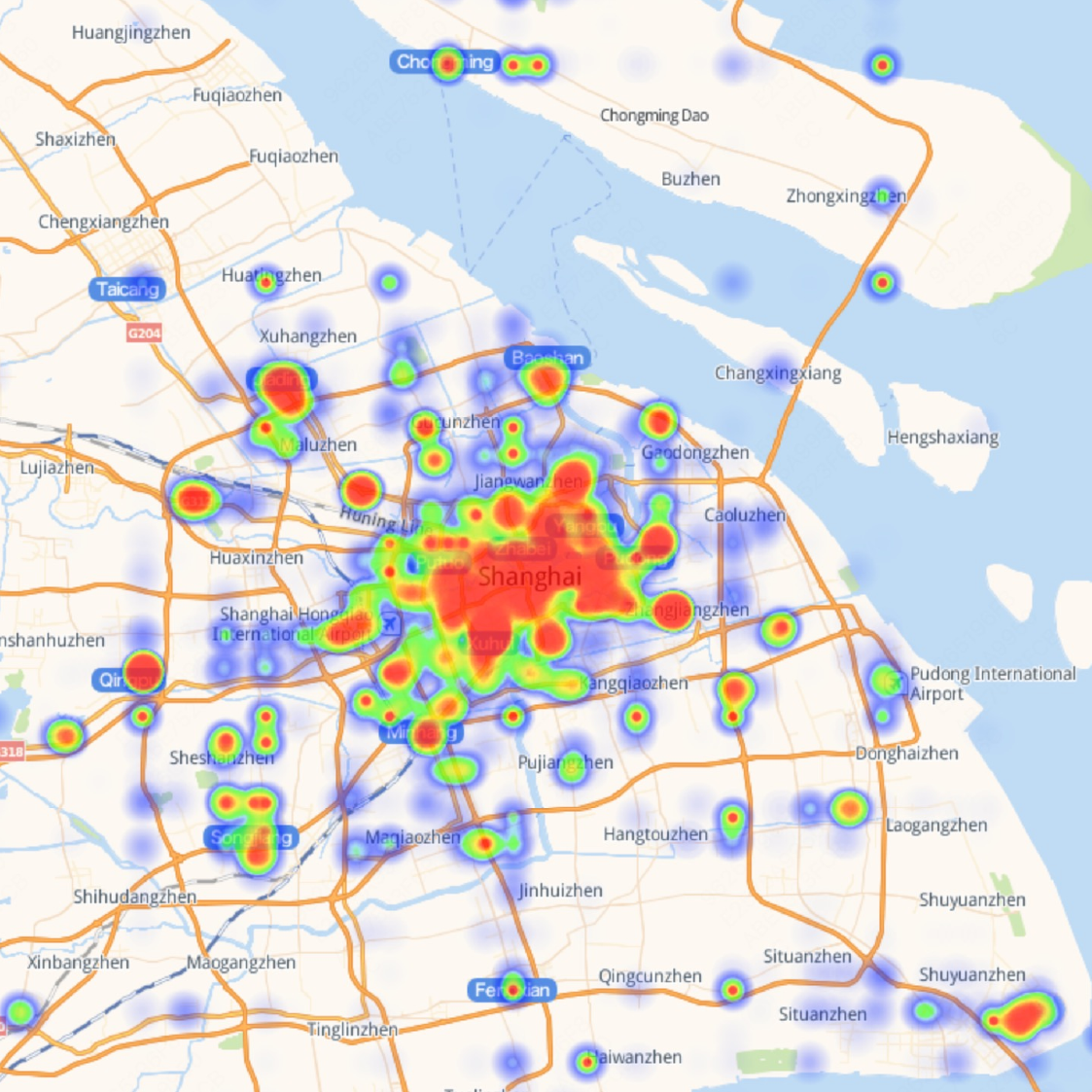}}
                \caption{Shanghai}
                \label{fig:shanghai_heatmap}
            \end{subfigure}%
            \hspace{0.015\columnwidth}
            \begin{subfigure}[b]{0.26\columnwidth}
                \centering
                \fbox{\includegraphics[width=\dimexpr\textwidth-2\fboxrule\relax, height=\dimexpr\textwidth-2\fboxrule\relax, keepaspectratio=false]{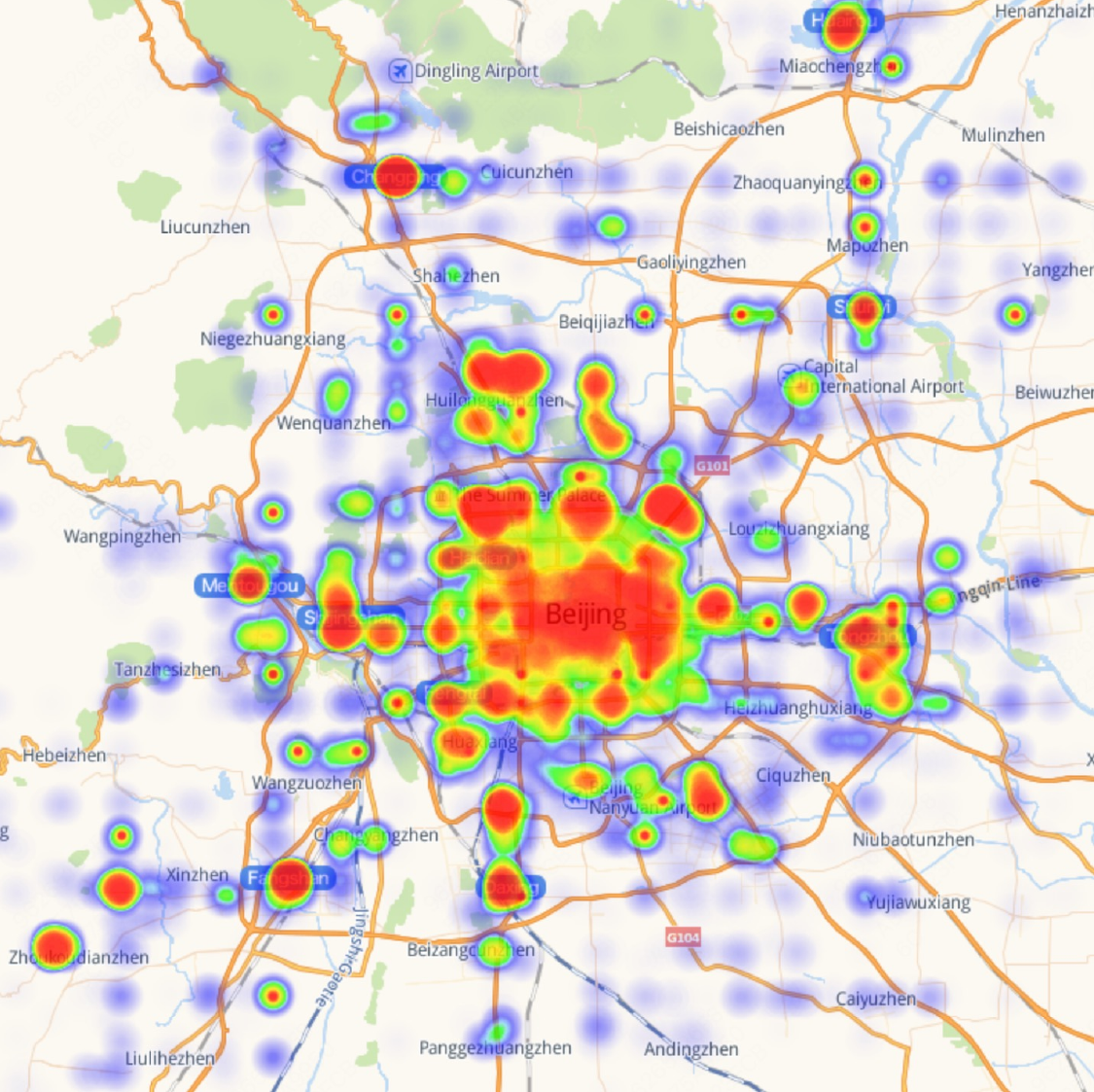}}
                \caption{Beijing}
                \label{fig:beijing_heatmap}
            \end{subfigure}%
            \hspace{0.015\columnwidth}
            \begin{subfigure}[b]{0.26\columnwidth}
                \centering
                \fbox{\includegraphics[width=\dimexpr\textwidth-2\fboxrule\relax, height=\dimexpr\textwidth-2\fboxrule\relax, keepaspectratio=false]{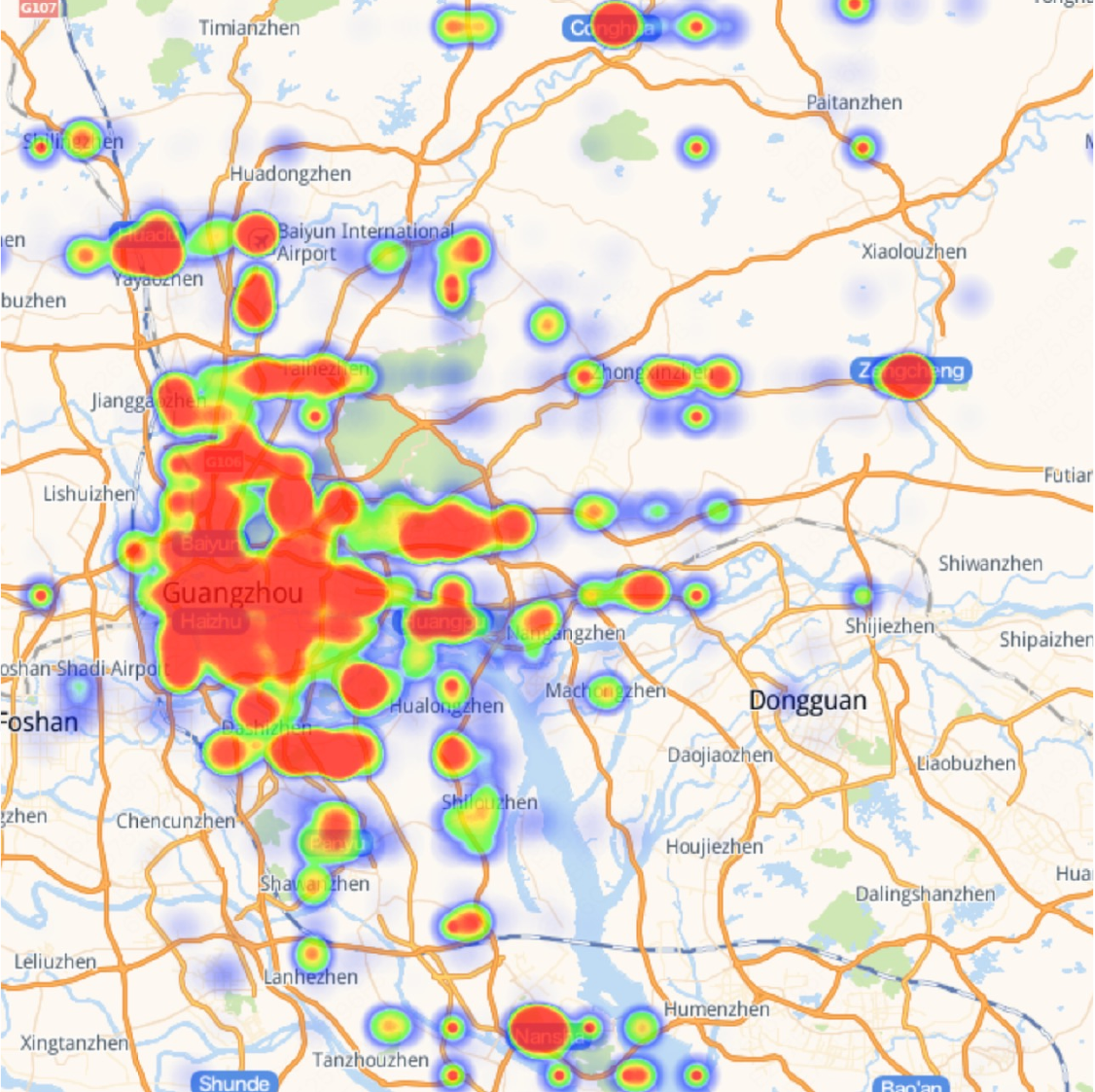}}
                \caption{Guangzhou}
                \label{fig:guangzhou_heatmap}
            \end{subfigure}
            \\[-0.2em]
            \begin{subfigure}[b]{0.26\columnwidth}
                \centering
                \fbox{\includegraphics[width=\dimexpr\textwidth-2\fboxrule\relax, height=\dimexpr\textwidth-2\fboxrule\relax, keepaspectratio=false]{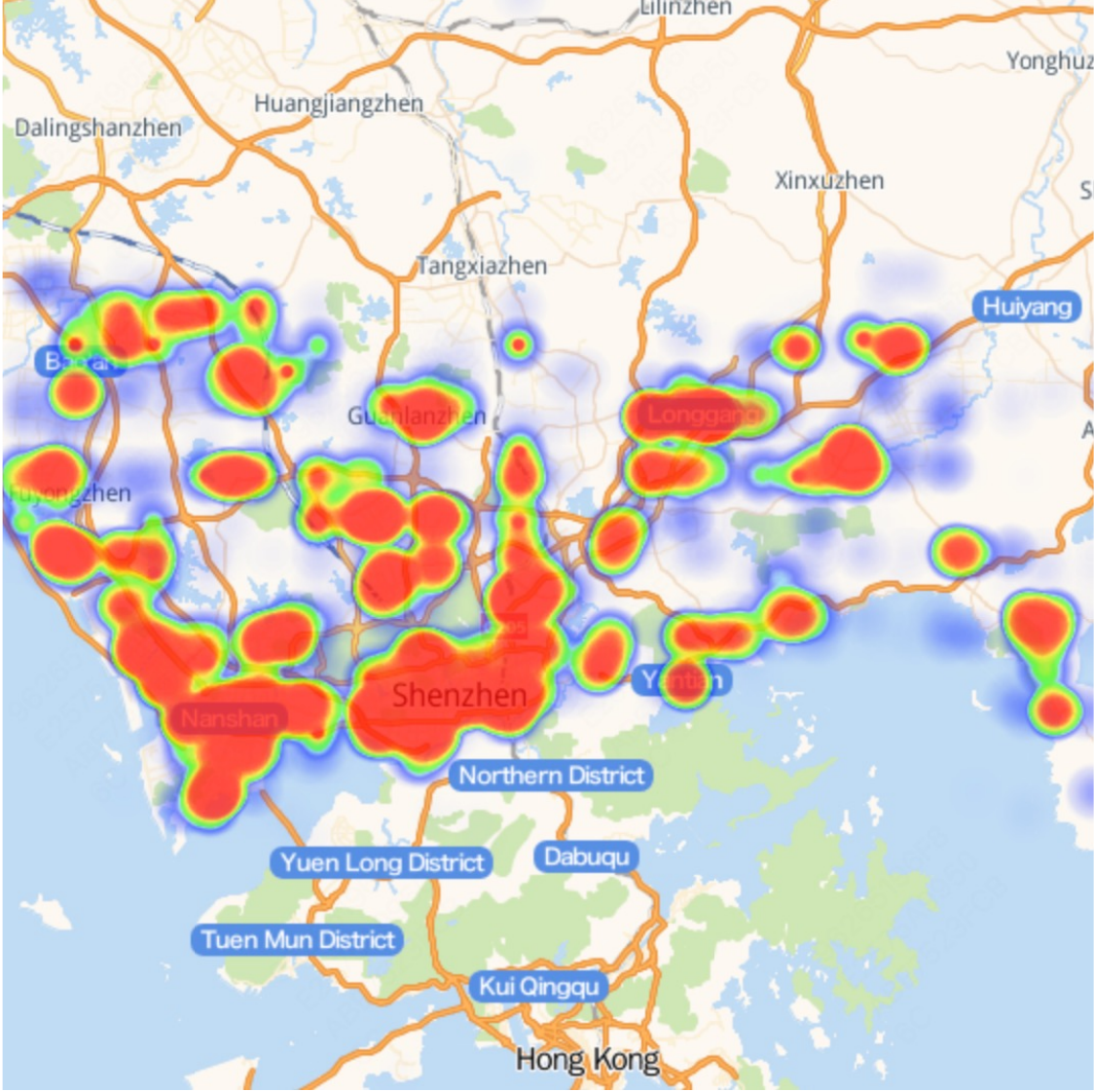}}
                \caption{Shenzhen}
                \label{fig:shenzhen_heatmap}
            \end{subfigure}%
            \hspace{0.015\columnwidth}
            \begin{subfigure}[b]{0.26\columnwidth}
                \centering
                \fbox{\includegraphics[width=\dimexpr\textwidth-2\fboxrule\relax, height=\dimexpr\textwidth-2\fboxrule\relax, keepaspectratio=false]{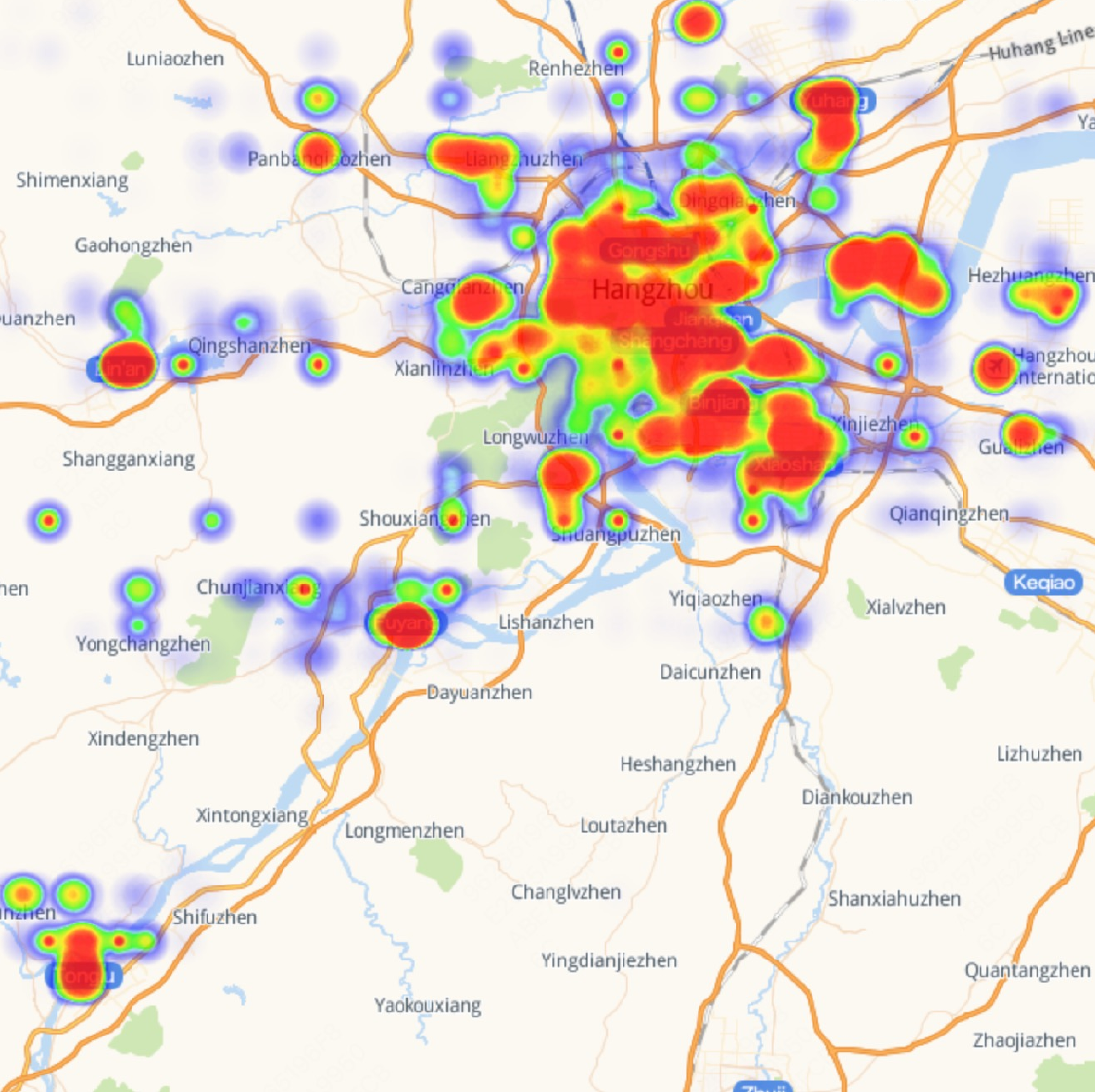}}
                \caption{Hangzhou}
                \label{fig:hangzhou_heatmap}
            \end{subfigure}
            \hspace{0.015\columnwidth}
            \begin{subfigure}[b]{0.26\columnwidth}
                \centering
                \fbox{\includegraphics[width=\dimexpr\textwidth-2\fboxrule\relax, height=\dimexpr\textwidth-2\fboxrule\relax, keepaspectratio=false]{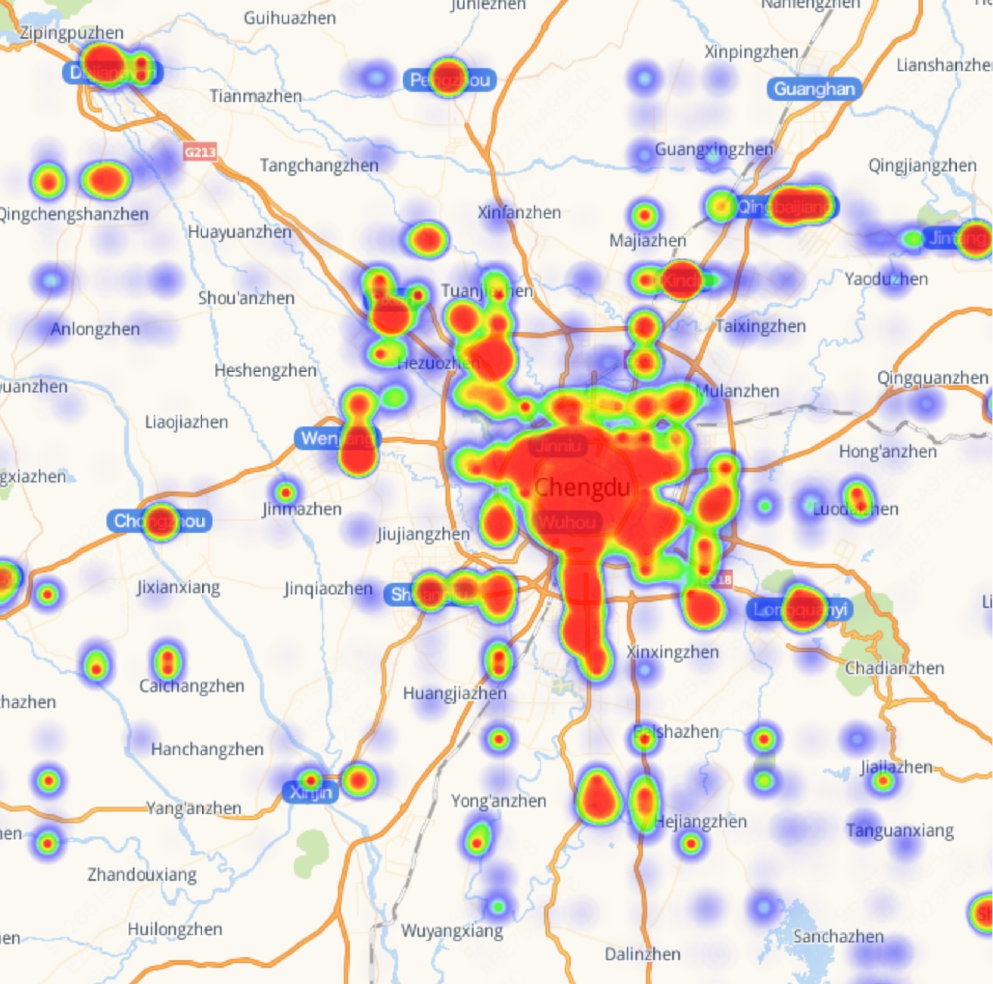}}
                \caption{Chengdu}
                \label{fig:chengdu_heatmap}
            \end{subfigure}%
            \\[-0.2em]
            \begin{subfigure}[b]{0.26\columnwidth}
                \centering
                \fbox{\includegraphics[width=\dimexpr\textwidth-2\fboxrule\relax, height=\dimexpr\textwidth-2\fboxrule\relax, keepaspectratio=false]{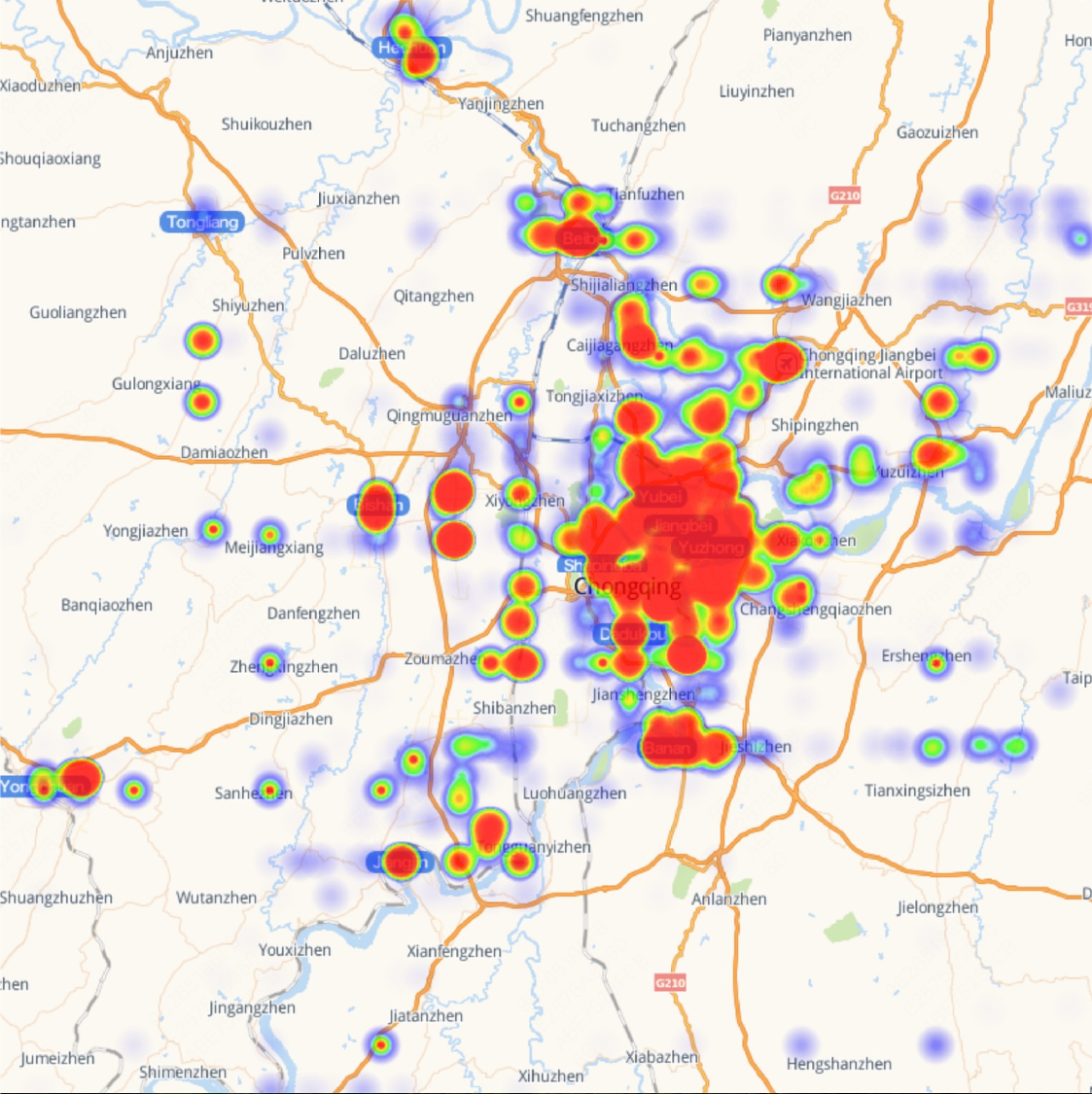}}
                \caption{Chongqing}
                \label{fig:chongqing_heatmap}
            \end{subfigure}%
            \hspace{0.015\columnwidth}
            \begin{subfigure}[b]{0.26\columnwidth}
                \centering
                \fbox{\includegraphics[width=\dimexpr\textwidth-2\fboxrule\relax, height=\dimexpr\textwidth-2\fboxrule\relax, keepaspectratio=false]{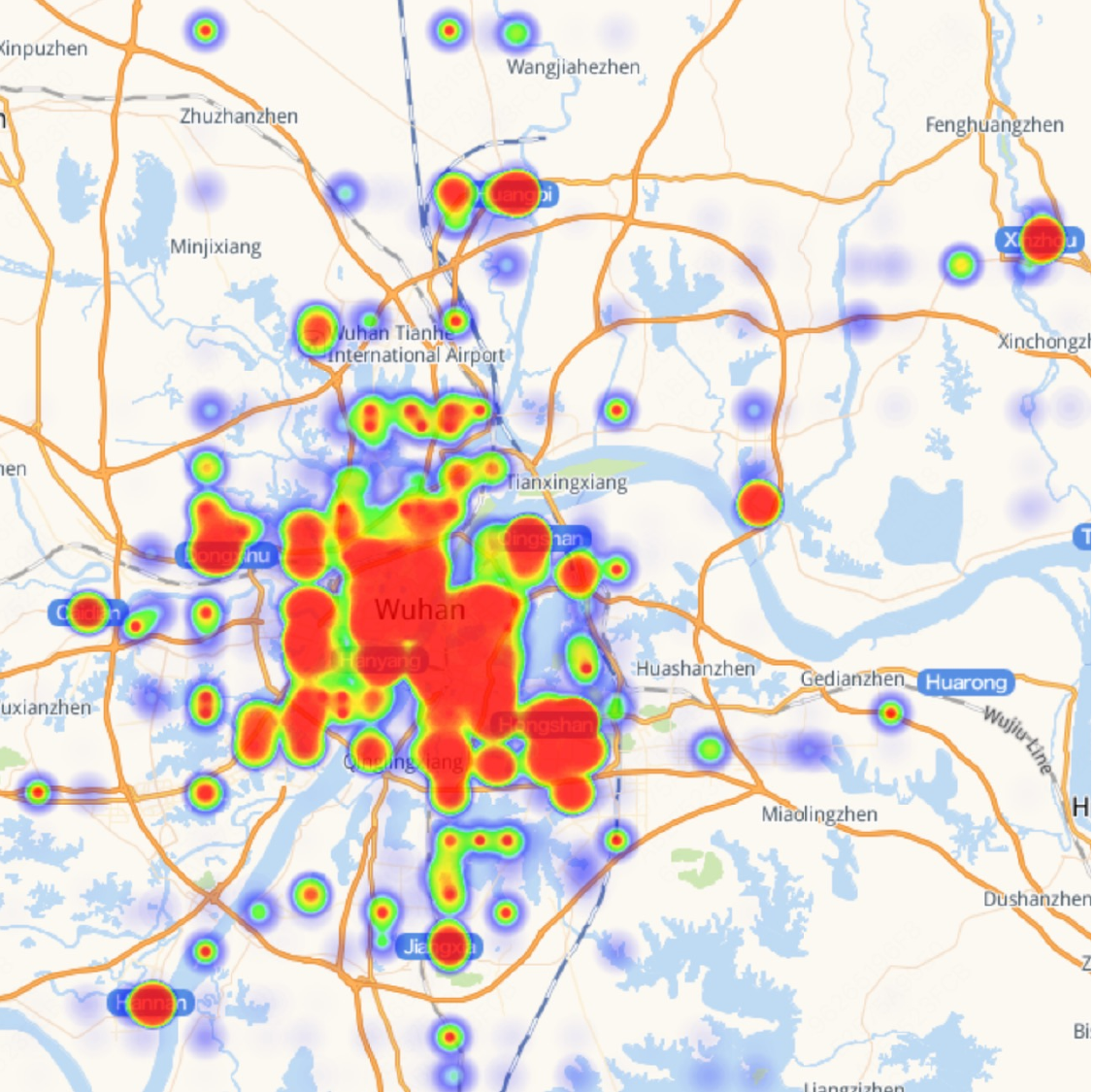}}
                \caption{Wuhan}
                \label{fig:wuhan_heatmap}
            \end{subfigure}%
            \hspace{0.015\columnwidth}
            \begin{subfigure}[b]{0.26\columnwidth}
                \centering
                \fbox{\includegraphics[width=\dimexpr\textwidth-2\fboxrule\relax, height=\dimexpr\textwidth-2\fboxrule\relax, keepaspectratio=false]{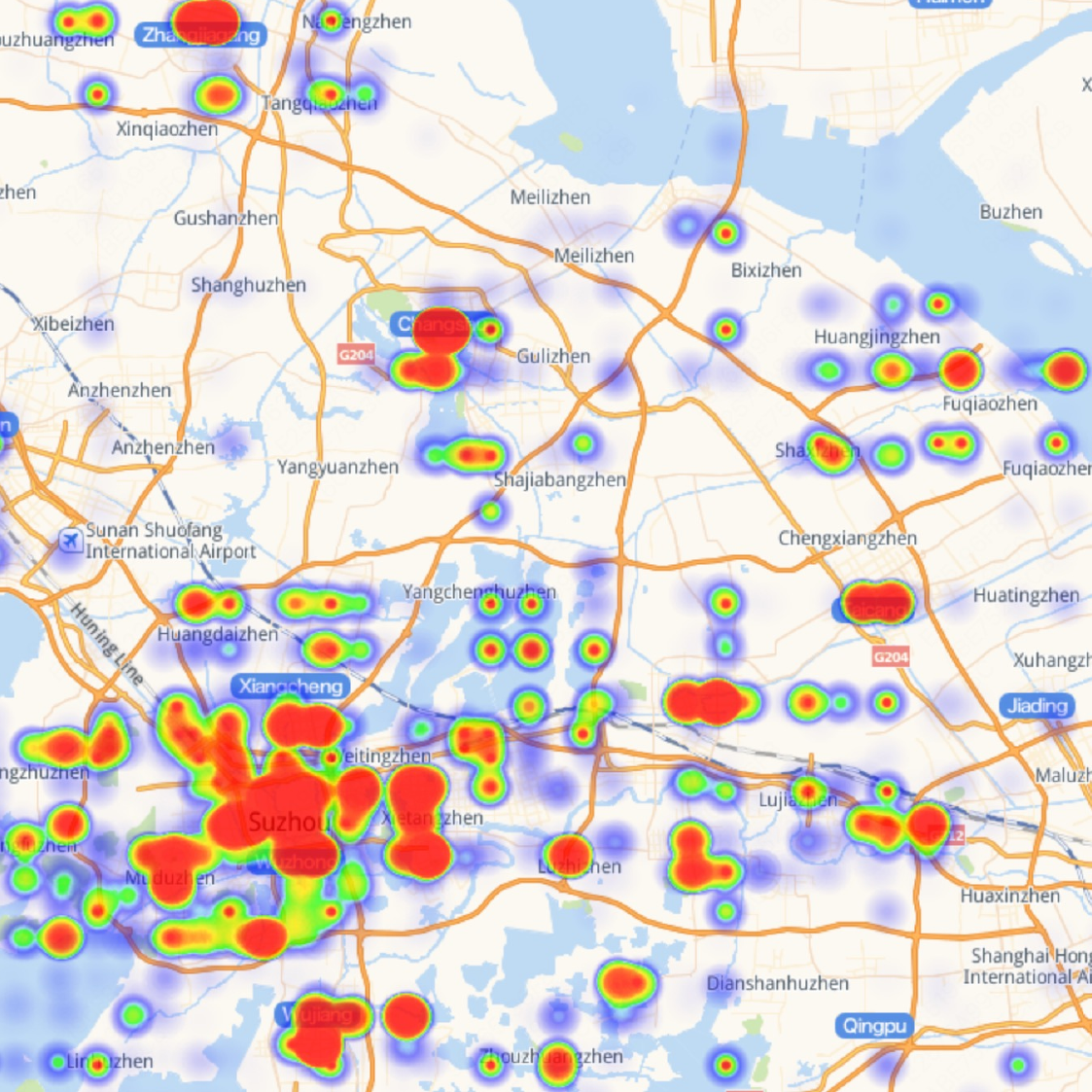}}
                \caption{Suzhou}
                \label{fig:suzhou_heatmap}
            \end{subfigure}%
        \end{tabular}
        &
        \includegraphics[height=0.25\columnwidth]{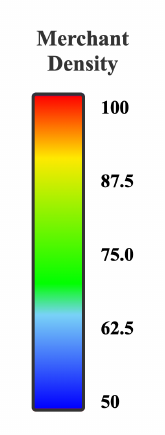}%
    \end{tabular}%
    }
    \vspace{-7pt}
    \caption{Geographical distribution heatmaps of 1,354,185 merchant data across 9 major cities in China. The color scale represents merchant density from 50 (blue) to 100 (red).}
    \label{fig:city_heatmaps}
    \vspace{-9pt}
\end{figure}

Overall, we construct 1,354,185 merchant profiles from over 1.6M original seed records through augmentation, privacy rewriting, and validation, forming the final \bench database. Category and city distributions align with raw data, preserving real-world distributions (Figure~\ref{fig:city_and_categories}). City heatmaps show realistic urban clusters (Figure~\ref{fig:city_heatmaps}).

\subsection{\rag Construction}
\label{sec:localrag}
To support efficient and accurate merchant retrieval for \emph{LRM}s, we build \textbf{\rag}, a Retrieval-Augmented Generation system on our merchant database (Figure~\ref{fig:evaluation}). \rag uses an embedding model to convert each merchant profile into a high-dimensional vector. Structured fields are directly encoded into vectors and indexed in a city-partitioned vector database. During query processing, the target city is first extracted from the query text, then the corresponding city-partitioned vector database is selected. The query is embedded with the \rag~embedding model and matched via cosine similarity against merchant vectors from that city, selecting the top-\textit{N} semantically related candidates through semantic retrieval.
\rag then applies a reranking model to the top-\textit{N} candidates, scoring query-merchant relevance and returning the top-\textit{K} merchants as the final retrieval results for LLM agents.

To evaluate retrieval quality, we employ two standard ranking metrics: Normalized Discounted Cumulative Gain (NDCG@$k$) and Mean Reciprocal Rank (MRR@$k$). NDCG@$k$ measures ranking quality by discounting relevance scores by position:
\begin{equation}
\small
\mathrm{NDCG@}k = \frac{1}{|Q|}\sum_{q \in Q} \frac{2^{y_{q,i}} - 1}{\log_2(1 + i)},
\end{equation}
where $|Q|$ is the number of queries and $y_{q,i}$ is the relevance score of the $i$-th document for query $q$. MRR@$k$ measures the average reciprocal rank of the first relevant document:
\begin{equation}
\small
\mathrm{MRR@}k = \frac{1}{|Q|}\sum_{q \in Q} \frac{1}{\mathrm{rank}_q},
\end{equation}
where $\mathrm{rank}_q$ is the position of the first relevant document for query $q$. Experimental results (\S~\ref{sec:rag_quality}) show that our two-stage pipeline outperforms single-stage retrieval on both metrics.

\subsection{Multi-hop QA Construction}
\label{sec:multi_hop_qa}

\begin{figure*}[]
    \centering
    \includegraphics[width=\linewidth]{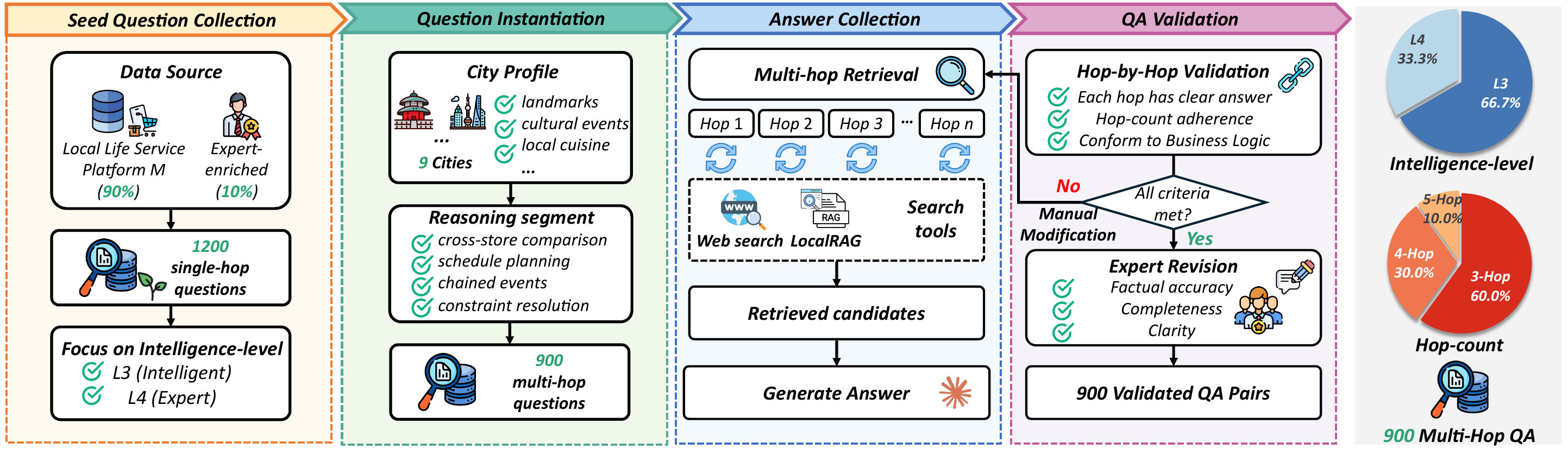}
    \caption{How Multi-hop QA pairs in \bench are constructed.
    }
    \label{fig:seed-question}
\end{figure*}
In this subsection, we present the multi-hop QA construction of \bench. Building upon the data construction pipeline described in the previous subsections, our multi-hop QA construction pipeline consists of four main stages: seed question collection, question instantiation, answer collection, and QA validation. Each of these stages is detailed in the following subsubsections.

\subsubsection{Seed Question Collection}
To ensure \bench reflects real-world local life service complexities, we conducted a statistical analysis to understand user search behaviors and service interactions. Based on requirement complexity and reasoning demands, we classify questions into five intelligence levels (Table~\ref{tab:intelligence_levels}): \textbf{L1 (Basic)} for simple factual queries; \textbf{L2 (Advanced)} for ambiguous or multi-condition searches; \textbf{L3 (Intelligent)} for composite requirements with multi-turn reasoning; \textbf{L4 (Expert)} for personalized planning; \textbf{L5 (AGI)} for cross-platform coordination with adaptive reflection.

We collect 1,200 single-hop questions from local life service platform M (90\%) and enriched sources (10\%) across 6 service categories (in \S~\ref{subsubsection:Multi-Scenario}). The platform data captures authentic user queries, while enriched data provides coverage for complex scenarios underrepresented in natural logs. Following the intelligence level taxonomy in Table~\ref{tab:intelligence_levels}, we focus on \textbf{L3 (Intelligent)} and \textbf{L4 (Expert)} questions, as they require multi-step reasoning within a static environment, while L5 (AGI) questions involve dynamic cross-platform interactions that are difficult to simulate offline.

\begin{table}[htbp]
\centering
\caption{Intelligence Level Classification for Questions.}
\label{tab:intelligence_levels}
\small
\renewcommand{\arraystretch}{1.0}
\resizebox{\linewidth}{!}{%
\begin{tabular}{c|c|c|c}
\toprule[1.5pt]
\rowcolor{gray!10}
\textbf{\makecell{Intellig-\\ence Level}} & \textbf{Grade} & \textbf{Requirement Understanding} & \textbf{Planning-Search-Reflection Loop} \\
\hline
\makecell{L1} & \makecell{Basic} & \begin{tabular}[c]{@{}c@{}}Standard expression of single \\precise requirement \end{tabular}& \begin{tabular}[c]{@{}c@{}}Supply matching search with $\leq$ \\2 conditions \end{tabular}\\
\hline
\rowcolor{gray!10}
L2 & \makecell{Advanced} & \begin{tabular}[c]{@{}c@{}}Ambiguous expression of single \\precise requirement; \\Standard expression of \\single ambiguous requirement\end{tabular} & \begin{tabular}[c]{@{}c@{}}Supply matching search with $\geq$ \\3 conditions;\\ Semantic retrieval search supply;\\ Multi-source information \\integration supply\end{tabular} \\
\hline
\makecell{L3} & \makecell{Intelligent} & \begin{tabular}[c]{@{}c@{}}Composite requirements;\\ Multi-turn conversational \\requirements\end{tabular} & \begin{tabular}[c]{@{}c@{}}Reasoning with certain complexity;\\ Semantic retrieval with external \\information dependency; \\Procedural planning requirements; \\Comparison and contrast scenarios\end{tabular} \\
\hline
\rowcolor{gray!10}
L4 & Expert & \begin{tabular}[c]{@{}c@{}}Personalized requirements\end{tabular} & \begin{tabular}[c]{@{}c@{}}External hard-to-access dependency; \\Open-ended planning requirements\end{tabular} \\
\hline
\makecell{L5} & \makecell{AGI} & \begin{tabular}[c]{@{}c@{}}Cross-platform coordination;\\ Complex decision-making;\\ Exception handling and reflection\end{tabular} & \begin{tabular}[c]{@{}c@{}}Cross-platform dependency;\\ Task execution;\\ Adaptive reflection\end{tabular} \\
\bottomrule[1.5pt]
\end{tabular}%
}
\end{table}


\subsubsection{Question Instantiation}
We instantiate the collected questions with city-specific information through a two-phase process. First, we construct city profiles for each of the 9 cities (\S~\ref{subsubsection:Multi-City Geographic}),  collecting structured information from public resources including landmarks, cultural events, local cuisine, price bands, etc. To facilitate consistent evaluation, temporal elements are anchored to static timestamps, creating a fixed reference point for all queries. The collected city data is preprocessed using automated scripts to remove duplicates, normalize formats, and correct inconsistencies, then validated by annotators to ensure factual accuracy and cultural authenticity. Second, annotators from platform M manually instantiate questions into concrete multi-hop queries. For each city, approximately 100 questions are selected from the collected 1,200 questions and grounded with city-specific information. To transform single-hop questions into multi-hop queries, annotators enrich questions with 2--4 reasoning segments based on business relevance and logical coherence. These segments include: cross-merchant comparison, schedule planning, chained events, and constraint resolution. We maintain target intelligence-level and hop-count distributions (Figure~\ref{fig:seed-question}), prioritizing underrepresented categories to ensure balanced coverage across service categories and cities in 
\S~\ref{subsubsection:Multi-Scenario}.

\subsubsection{Answer Collection}
\label{subsection:answer collection}

For each instantiated question, we generate reference answers through a multi-step process using \rag (\S~\ref{sec:localrag}), web search, and LLM-based generation. We first check whether the question needs real-time information (e.g., current events, weather, news) or can be answered with merchant data alone. If real-time signals are present, we invoke web search; otherwise, the query relies on the city-partitioned merchant vector database. For all questions, \rag retrieves merchants scoped to the target city. For multi-hop questions, the hop count, retrieval target, and required search tool for each hop are predetermined during question construction. Search tools then execute iterative retrieval per hop, gathering merchants across categories/locations and combining them with any web-search snippets to form the full reasoning context. 
Claude Opus 4.5~\cite{anthropic2025claude41} then generate preliminary answers with expert oversight. The LLM reranks each hop's retrieved candidates and selects the top-$k$ passages ($k{=}5$) as evidence, then synthesizes a final answer grounded in these snippets while following the fixed hop plan and the original query. The prompt enforces: (1) grounding in retrieved evidence, (2) alignment with the fixed hop plan, (3) explicit citation of merchant and web evidence, and (4) avoidance of unsupported content. The resulting question-answer pairs serve as input for validation.

\subsubsection{QA Validation}
\label{lab:data validation}

All preliminary QA pairs undergo a two-stage validation process (Figure~\ref{fig:seed-question}). First, hop-by-hop validation ensures: (1) every hop has a clear answer, meaning retrieval targets must be found through \rag or web search with sufficient evidence; (2) strict adherence to the target hop-count chain; (3) conformity to local life service business logic. Questions failing these criteria are manually modified and re-submitted to answer collection (\S~\ref{subsection:answer collection}). Second, three domain experts independently review answers for factual accuracy, completeness, and clarity, resolving disagreements through consensus. Answers are deterministic with balanced distribution across service categories and cities (\S~\ref{lab:Seed Data Collection}), yielding 900 high-quality QA pairs for the final benchmark (Figure~\ref{fig:seed-question}).

\begin{figure*}[t]
    \centering
    \includegraphics[width=\linewidth]{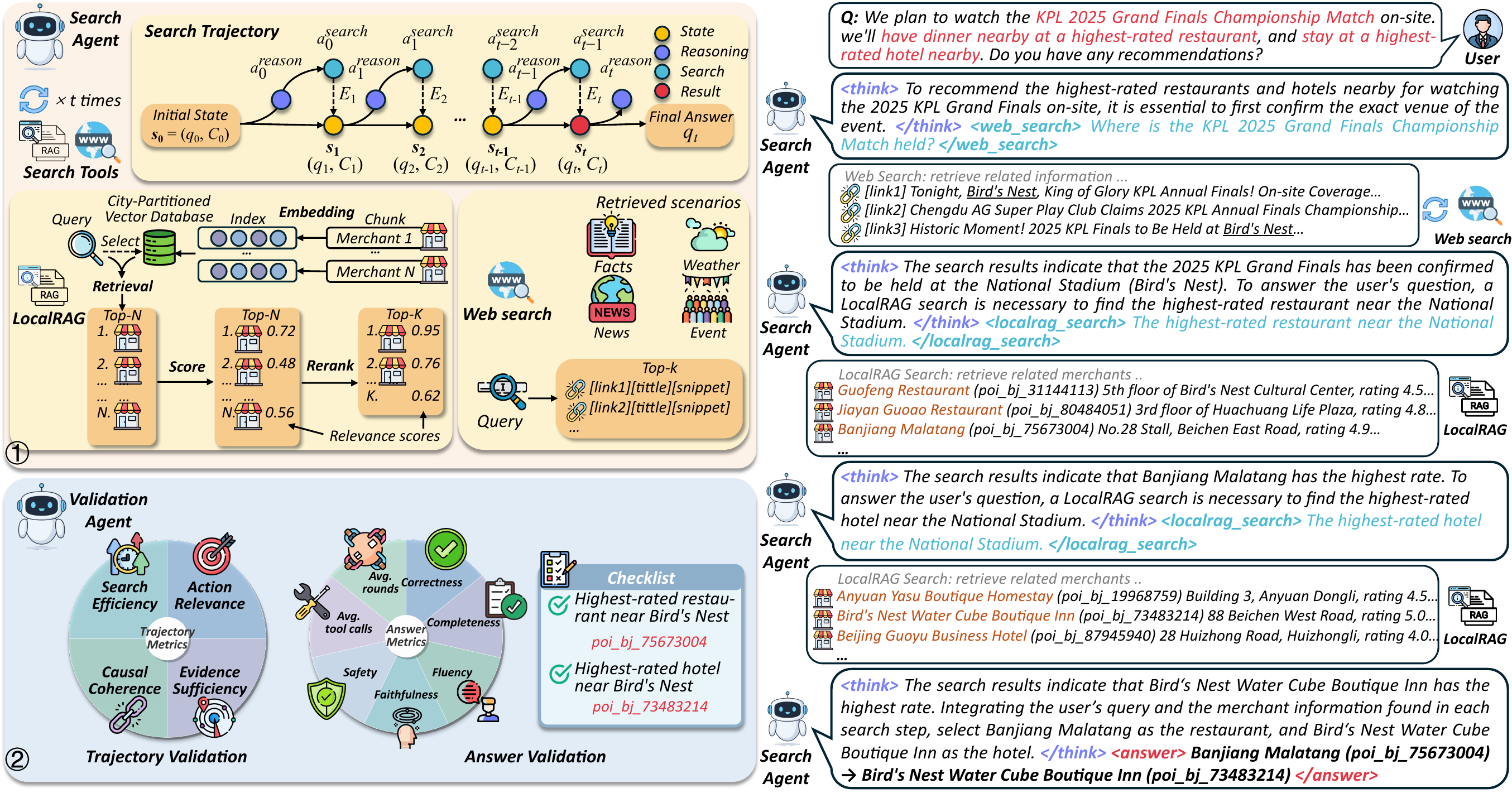}
    \caption{Overview of \eval. 
    \ding{182} Search Agent follows ReAct structure and performs iterative multi-hop reasoning using \rag and Web Search; \ding{183} Validation Agent assesses answer and trajectory quality. 
    }
    \label{fig:evaluation}
\end{figure*}

\section{\eval}
We design \eval for automated testing with \bench. It  contains a \emph{Search Agent} for multi-hop retrieval and answering, and a \emph{Validation Agent} for quality examination of answer and reasoning trajectory (Figure~\ref{fig:evaluation}).

\textbf{Search Agent.} The Search Agent uses two tools: \rag for vector-DB merchant retrieval and web search for real-time information (facts, weather, news, events) via Baidu APIs~\cite{liu2021pre}. It performs iterative search up to \emph{N} rounds, calling each tool at most once per round to refine answers. For each query $q_0$, it records the trajectory $\tau=\{s_0,a_0,s_1,a_1,\dots,s_t\}$, with states $s_i=(q_i,C_i)$ storing the current sub-question and accumulated evidence. At step $i$, it selects a reasoning action $a_i^{\text{reason}}$ and, if needed, a search action $a_i^{\text{search}}$ to obtain evidence $E_{i+1}$, update $C_{i+1}=C_i\cup E_{i+1}$, and generate $q_{i+1}$ from $(q_i,C_{i+1})$. A final reasoning step synthesizes $q_t$, and $s_{i+1}=F(s_i,a_i)$ forms a directed chain over sub-questions and evidence for reasoning.

\textbf{Validation Agent.} We employ an LLM-as-judge (Claude Opus 4.5) to evaluate each agent interaction along two dimensions: \emph{answer quality} and \emph{trajectory quality}. \textit{Answer quality} is assessed on 7 metrics: (1) \textbf{correctness} is verified by checking whether the retrieved merchant satisfies all constraints defined in the query's human-constructed multi-attribute \textit{checklist} (if multiple merchants meet all listed conditions, each  correct); (2) \textbf{completeness} evaluates whether the answer fully addresses all aspects of the query; (3) \textbf{fluency} assesses the readability and coherence of the generated text; (4) \textbf{faithfulness} checks for hallucination by ensuring all claims are supported by retrieved evidence; (5) \textbf{safety} identifies potentially harmful or inappropriate content; (6) \textbf{avg. tool calls} and (7) \textbf{avg. rounds} are rule-based efficiency metrics that count the number of tool calls and search rounds, respectively.  
\textit{Trajectory quality} is scored on 4 dimensions: (1) \textbf{action relevance} measures whether each search step aligns with the intended information need; (2) \textbf{evidence sufficiency} assesses whether the collected evidence is adequate to answer the query; (3) \textbf{causal coherence} verifies that later reasoning steps correctly reference and build upon earlier evidence; (4) \textbf{search efficiency} penalizes redundant or ineffective tool calls that do not contribute to the final answer.
Each instance is evaluated 5 times per dimension and averaged. Systems are anonymized, question order is randomized, and the judge model is distinct from all evaluated models.

\begin{table*}[t]
    \centering
    \setlength{\tabcolsep}{1.5pt}  
    \caption{Performance comparison of different models on \bench (Max Round \emph{N} = 5). The top and worst performing results are highlighted in \colorbox{backred!50}{red} (1\textsuperscript{st}) and \colorbox{backblue!75}{blue} (bottom) backgrounds, respectively.}
    \vspace{-5pt}
    \label{tab:model_performance_full}
    \small
    \resizebox{\linewidth}{!}{
    \begin{tabular}{@{}l|ccccccc|cccc@{}}
    \toprule[1.5pt]
    \multirow{2}{*}{\textbf{Models}}
    & \multicolumn{7}{c|}{\textbf{Answer Evaluation Metrics}}
    & \multicolumn{4}{c}{\textbf{Trajectory Evaluation Metrics}} \\ \cline{2-12}
    & \textbf{\makecell{Avg. tool\\calls}}
    & \textbf{\makecell{Avg.\\rounds}}
    & \textbf{\makecell{Correct-\\ness(\%)}}
    & \textbf{\makecell{Complete-\\ness(\%)}}
    & \textbf{\makecell{Fluency\\(\%)}}
    & \textbf{\makecell{Faithful-\\ness(\%)}}
    & \textbf{\makecell{Safety\\(\%)}}
    & \textbf{\makecell{Action\\Relevance (\%)}}
    & \textbf{\makecell{Evidence\\Sufficiency (\%)}}
    & \textbf{\makecell{Causal\\Coherence(\%)}}
    & \textbf{\makecell{Search\\Efficiency (\%)}}
    \\[3pt]
    \midrule
    \textbf{Qwen3-235B-A22B} (w/o thinking) & 2.00 & 2.93 & 21.18 & 50.94 & 69.16 & 25.28 & 79.72 & 75.22 & 43.61 & 50.68 & 45.99 \\
    \textbf{Qwen3-235B-A22B} (w/ thinking) & 2.31 & 3.17 & 30.24 & 71.20 & \colorbox{backred!50}{\textbf{\underline{71.58}}} & 26.90 & 81.76 & 80.42 & 45.75 & 52.04 & 48.63 \\\hline
    \textbf{Qwen3-32B} (w/o thinking)  & 2.78 & 3.11 & 19.76 & 40.96 & 68.50 & \colorbox{backblue!75}{21.38} & 80.76 & 74.67 & 46.52 & 48.82 & 49.84 \\
    \textbf{Qwen3-32B} (w/ thinking)  & 2.80 & 3.12 & 25.63 & 40.66 & 68.44 & 22.40 & 79.54 & 75.06 & 46.99 & 48.87 & 49.13 \\\hline
    \textbf{Qwen3-14B} (w/o thinking)  & 2.53 & \colorbox{backblue!75}{2.07} & 24.21 & \colorbox{backblue!75}{40.60} & 69.62 & 27.44 & 80.78 & 80.46 & 46.44 & 50.96 & 50.02 \\
    \textbf{Qwen3-14B} (w/ thinking)  & 2.57 & 2.12 & 25.17 & 40.98 & 69.32 & 28.40 & 80.44 & \colorbox{backred!50}{\textbf{\underline{81.19}}} & 47.24 & 52.52 & 51.65 \\\hline
    \textbf{GPT-4.1} & \colorbox{backblue!75}{1.73} & 2.42 & 18.47 & 45.37 & \colorbox{backblue!75}{65.93} & 28.76 & \colorbox{backblue!75}{77.38} & \colorbox{backblue!75}{68.47} & \colorbox{backblue!75}{38.62} & \colorbox{backblue!75}{45.83} & 42.29 \\
    \textbf{o3(high)} & 2.91 & 3.38 & 31.53 & 69.64 & 70.72 & 33.89 & 81.87 & 75.93 & 44.71 & 51.78 & 42.96 \\\hline
    \textbf{Gemini-2.5-Flash} & 1.84 & 2.51 & 20.97 & 58.44 & 68.04 & 35.72 & 79.70 & 70.58 & 41.61 & 48.94 & 46.91 \\
    \textbf{Gemini-2.5-Pro}   & 2.75 & 3.10 & 32.34 & 71.03 & 71.12 & 34.93 & 82.32 & 77.31 & 45.73 & 52.87 & \colorbox{backblue!75}{41.78} \\\hline
    \textbf{LongCat-Flash-Chat} & 2.34 & 3.07 & 25.28 & 52.98 & 69.45 & 27.49 & \colorbox{backred!50}{\textbf{\underline{83.61}}} & 77.78 & 47.33 & 50.86 & 52.29 \\
    \textbf{LongCat-Flash-Thinking} & 3.04 & 3.20 & 30.68 & 68.83 & 69.07 & 31.47 & 80.10 & 78.50 & 47.37 & \colorbox{backred!50}{\textbf{\underline{53.18}}} & 53.27 \\\hline
    \textbf{GLM-4.6} (w/o thinking) & 2.86 & 3.86 & 28.97 & 76.45 & 70.37 & 35.40 & 81.40 & 74.28 & 48.44 & 50.79 & 52.76 \\
    \textbf{GLM-4.6} (w/ thinking) & 3.08 & 4.06 & 32.83 & 76.83 & 70.27 & 37.48 & 81.30 & 77.66 & \colorbox{backred!50}{\textbf{\underline{48.90}}} & 52.67 & 54.43 \\\hline
    \textbf{DeepSeek-V3.2} (w/o thinking) & 3.12 & 4.11 & 32.74 & 77.08 & 70.12 & 35.96 & 81.52 & 75.51 & 48.49 & 52.23 & 54.33 \\
    \textbf{DeepSeek-V3.2} (w/ thinking) & \colorbox{backred!50}{\textbf{\underline{3.21}}} & \colorbox{backred!50}{\textbf{\underline{4.20}}} & \colorbox{backred!50}{\textbf{\underline{35.60}}} & \colorbox{backred!50}{\textbf{\underline{77.56}}} & 70.92 & \colorbox{backred!50}{\textbf{\underline{39.78}}} & 81.13 & 75.58 & 48.86 & 52.62 & \colorbox{backred!50}{\textbf{\underline{54.83}}} \\

    \bottomrule[1.5pt]
    \end{tabular}
    }
    \vspace{-5pt}
    \end{table*}

\begin{table*}[t]
    \centering
    \setlength{\tabcolsep}{3pt}  
        \caption{Ablation study of tool integration on thinking models using \bench (Max Round \emph{N} = 5). The top and worst performing results are highlighted in \colorbox{backred!50}{red} (1\textsuperscript{st}) and \colorbox{backblue!75}{blue} (bottom) backgrounds, respectively.}
    \vspace{-5pt}
    \label{tab:model_performance}
    \small
    \resizebox{\linewidth}{!}{
    \begin{tabular}{@{}l|ccccccc|cccc@{}}
    \toprule[1.5pt]
    \multirow{2}{*}{\textbf{Methods}}
    & \multicolumn{7}{c|}{\textbf{Answer Evaluation Metrics}}
    & \multicolumn{4}{c}{\textbf{Trajectory Evaluation Metrics}} \\ \cline{2-12}
    & \textbf{\makecell{Avg. tool\\calls}}
    & \textbf{\makecell{Avg.\\rounds}}
    & \textbf{\makecell{Correct-\\ness(\%)}}
    & \textbf{\makecell{Complete-\\ness(\%)}}
    & \textbf{\makecell{Fluency\\(\%)}}
    & \textbf{\makecell{Faithful-\\ness(\%)}}
    & \textbf{\makecell{Safety\\(\%)}}
    & \textbf{\makecell{Action\\Relevance (\%)}}
    & \textbf{\makecell{Evidence\\Sufficiency (\%)}}
    & \textbf{\makecell{Causal\\Coherence(\%)}}
    & \textbf{\makecell{Search\\Efficiency (\%)}}
    \\
        \midrule
        \textbf{\rag} workflow & 1.00 & \colorbox{backblue!75}{1.00} & 15.22 & 58.30 & 70.80 & \colorbox{backred!50}{\textbf{\underline{66.10}}} & 85.40 & 37.70 & 23.53 & 32.77 & 39.35
    \\\hline
        \textbf{Gemini-2.5-Pro} (w/o tools)   & \colorbox{backblue!75}{0.00} & \colorbox{backblue!75}{1.00} & \colorbox{backblue!75}{6.33\textsuperscript{*}} & 38.42 & 91.65 & \colorbox{backblue!75}{9.82} & \colorbox{backred!50}{\textbf{\underline{99.89}}} & \colorbox{backblue!75}{25.21} & 8.72 & \colorbox{backblue!75}{21.70} & \colorbox{backblue!75}{13.62} \\
    + \rag   & 1.86 & 2.83 & 23.11 & 63.92 & 82.32 & 38.65 & 84.95 & 68.10 & 37.37 & 46.93 & 39.96 \\
        + \rag \& \emph{web search}   & 2.75 & 3.10 & 32.34 & 71.03 & 71.12 & 34.93 & 82.32  & 77.31 & 45.73 & \colorbox{backred!50}{\textbf{\underline{52.87}}} & 41.78  \\\hline
        \textbf{GLM-4.6} (w/o tools) & \colorbox{backblue!75}{0.00} & \colorbox{backblue!75}{1.00} & {7.89\textsuperscript{*}} & \colorbox{backblue!75}{36.06} & \colorbox{backred!50}{\textbf{\underline{91.92}}} & 12.12 & 99.67 & 46.88 & \colorbox{backblue!75}{7.92} & 28.85 & 17.08  \\
    + \rag & 2.33 & 3.32 & 23.89 & 69.21 & 83.75 & 41.45 & 84.72 & 63.33 & 36.21 & 48.02 & 52.05 \\
        + \rag \& \emph{web search} & 3.08 & 4.06 & 32.83 & 76.83 & \colorbox{backblue!75}{70.27} & 37.48 & 81.30 & \colorbox{backred!50}{\textbf{\underline{77.66}}} & \colorbox{backred!50}{\textbf{\underline{48.90}}} & 52.67 & 54.43  \\\hline
        \textbf{DeepSeek-V3.2} (w/o tools) & \colorbox{backblue!75}{0.00} & \colorbox{backblue!75}{1.00} & {9.78\textsuperscript{*}} & 39.30 & 90.20 & 18.56 & 99.55 & 52.65 & 10.80 & 32.10 & 41.15 \\
    + \rag & 2.88 & 3.87 & 27.00 & 70.05 & 82.98 & 40.59 & 83.85  & 65.55 & 42.78 & 47.85 & 54.07 \\
        + \rag \& \emph{web search} & \colorbox{backred!50}{\textbf{\underline{3.21}}} & \colorbox{backred!50}{\textbf{\underline{4.20}}} & \colorbox{backred!50}{\textbf{\underline{35.60}}} & \colorbox{backred!50}{\textbf{\underline{77.56}}} & 70.92 & 39.78 & \colorbox{backblue!75}{81.13} & 75.58 & 48.86 & 52.62 & \colorbox{backred!50}{\textbf{\underline{54.83}}}  \\
    \bottomrule[1.5pt]
    \end{tabular}
    }
\begin{flushleft}\footnotesize{* \textbf{W/o tools}:  A result is correct if merchant names and attributes match Platform M's ground truth. }\end{flushleft}

    \vspace{-5pt}
    \end{table*}

\begin{table*}[t]
\centering
\caption{Sensitivity analysis of Max Round \emph{N} on model performance. The top and worst performing results are highlighted in \colorbox{backred!50}{red} (1\textsuperscript{st}) and \colorbox{backblue!75}{blue} (bottom) backgrounds, respectively.}
\vspace{-5pt}
\label{tab:sensitivity}
\resizebox{\linewidth}{!}{
\begin{tabular}{c|ccccccccccc}
\toprule[1.5pt]
\multirow{2}{*}{\textbf{\makecell{Max\\Round}}}
& \multicolumn{7}{c}{\textbf{Answer Evaluation Metrics}}
& \multicolumn{4}{c}{\textbf{Trajectory Evaluation Metrics}} \\ \cline{2-12}
& \textbf{\makecell{Avg. tool\\calls}}
& \textbf{\makecell{Avg.\\rounds}}
& \textbf{\makecell{Correct-\\ness(\%)}}
& \textbf{\makecell{Complete-\\ness(\%)}}
& \textbf{\makecell{Fluency\\(\%)}}
& \textbf{\makecell{Faithful-\\ness(\%)}}
& \textbf{\makecell{Safety\\(\%)}}
& \textbf{\makecell{Action\\Relevance (\%)}}
& \textbf{\makecell{Evidence\\Sufficiency (\%)}}
& \textbf{\makecell{Causal\\Coherence(\%)}}
& \textbf{\makecell{Search\\Efficiency (\%)}}
\\ \hline
$N = 2$ & \colorbox{backblue!75}{2.46} & \colorbox{backblue!75}{2.82} & \colorbox{backblue!75}{30.56} & \colorbox{backblue!75}{70.61} & \colorbox{backred!50}{\textbf{\underline{76.44}}} & \colorbox{backred!50}{\textbf{\underline{54.65}}} & \colorbox{backred!50}{\textbf{\underline{89.94}}} & \colorbox{backblue!75}{59.94} & \colorbox{backblue!75}{31.17} & \colorbox{backblue!75}{44.82} & \colorbox{backblue!75}{40.77} \\
$N = 3$ & 2.78 & 3.30 & 33.56 & 74.79 & 73.57 & 43.50 & 84.58 & 61.68 & 42.96 & 47.42 & 48.50 \\
$N = 4$ & 3.03 & 3.95 & 34.22 & 76.34 & 72.27 & 41.26 & 82.53 & 72.97 & 49.28 & 51.32 & 54.13 \\
$N = 5$ & 3.21 & 4.20 & \colorbox{backred!50}{\textbf{\underline{35.60}}} & \colorbox{backred!50}{\textbf{\underline{77.56}}} & 70.92 & 39.78 & 81.13 & \colorbox{backred!50}{\textbf{\underline{75.58}}} & 48.86 & \colorbox{backred!50}{\textbf{\underline{52.62}}} & \colorbox{backred!50}{\textbf{\underline{54.83}}} \\
$N = 6$ & 3.76 & 4.75 & 34.89 & 74.52 & 70.36 & 38.56 & 80.91 & 68.63 & 49.28 & 51.65 & 47.10 \\
$N = 7$ & 3.87 & 4.85 & 33.67 & 73.60 & \colorbox{backblue!75}{70.19} & 37.36 & 80.84 & 69.50 & 49.37 & 48.20 & 44.99 \\
$N = 8$ & \colorbox{backred!50}{\textbf{\underline{4.02}}} & \colorbox{backred!50}{\textbf{\underline{5.01}}} & 32.44 & 73.15 & 70.24 & \colorbox{backblue!75}{37.35} & \colorbox{backblue!75}{80.39} & 66.89 & \colorbox{backred!50}{\textbf{\underline{49.39}}} & 47.94 & 43.58 \\
\bottomrule[1.5pt]
\end{tabular}
}
\vspace{-7pt}
\end{table*}
\vspace{-5pt}
\section{Evaluation}
\label{sec:evaluation}
\subsection{Evaluated Models}
We evaluate 16 state-of-the-art thinking and non-thinking models for agents: 
GPT series (GPT-4.1, o3 (high))~\cite{openai2025gpt4-1}, 
Gemini series (Gemini-2.5-Flash, Gemini-2.5-Pro)~\cite{comanici2025gemini}, LongCat series (LongCat-Flash-Chat, LongCat-Flash-Thinking)~\cite{team2025longcat}, DeepSeek-V3.2~\cite{deepseek2025v3}, GLM-4.6~\cite{zeng2025glm}, and Qwen3 series (Qwen3-14B, Qwen3-32B, Qwen3-235B-A22B)~\cite{yang2025qwen3}. Unless specified, \eval uses default settings: maximum tool call rounds \emph{N} = 5; \rag with Qwen3-Embedding-8B and Qwen3-Reranker-8B~\cite{zhang2025qwen3} (top-\emph{N} = 100 dense retrieval, top-\emph{K} = 20 reranking); web search top-\emph{k} = 20; and LLM temperature = 0. Experiments run on Intel Xeon Gold 5218 @ 2.30GHz with 1 $\times$ H20-141G GPU.

\subsection{Data and \rag Quality }
\label{sec:rag_quality}
We evaluate \bench database with the two-stage framework in \S\ref{sec:data_validation}, sampling 1,000 merchants. 
As shown in Table~\ref{tab:evaluation_comparison}, the augmented and privacy-rewritten data achieve high overall quality scores of 0.86 and 0.92 (out of 1).

We quantify human-machine agreement on the 20 fields by computing weighted Cohen's $\kappa$ ~\cite{cohen1960coefficient} per field $f$ against each of three evaluators and averaging over evaluators and fields.
\begin{equation}
\small
\bar{\kappa} = \frac{1}{F}\sum_{f=1}^{F}\left(\frac{1}{3}\sum_{i=1}^{3} \kappa_{f,i}\right).
\end{equation}
The same 3 evaluators run blind 5-point Likert checks on consistency, completeness, and anonymization (4$\sim$5 = satisfied, 1$\sim$3 = not satisfied). The model reaches $\bar{\kappa}=0.74$ (95\% CI: $[0.71, 0.77]$), corresponding to 88.34\% raw agreement.

Since retrieval quality directly affects downstream QA performance, we evaluate \rag on 27,360 hop-level queries using NDCG@10/20 and MRR@10/20 over the top-20 results of the full retrieval pipeline (\S\ref{sec:localrag}). Our best configuration, Qwen3-Embedding-8B, reaches NDCG@10/20 of 0.84/0.87 and MRR@10/20 of 0.82/0.83.
These hop-level queries come from the multi-hop QA construction in \S\ref{sec:multi_hop_qa} with ground-truth evidence for each hop.


\subsection{Overall Performance}
The choice of underlying \emph{LRM} significantly influences performance across all metrics. \textbf{DeepSeek-V3.2} achieves the highest Correctness (35.60\%), followed by GLM-4.6 (32.83\%) and Gemini-2.5-Pro (32.34\%), and also leads in Completeness (77.56\%) and Faithfulness (39.78\%).
Thinking models outperform non-thinking ones in answer quality metrics (Correctness: 30.49\% vs. 21.30\%, Completeness: 64.59\% vs. 49.10\%) and most trajectory evaluation metrics, with similar Search Efficiency. Despite these variations, overall performance remains limited, with the best model achieving only 35.60\% Correctness, underscoring the substantial challenge of multi-hop local search. Across all 16 models, correctness further drops on L4 tasks (25.10\%) compared to L3 (30.57\%), reflecting the greater challenge of L4.

\textbf{Human evaluation.}
While LLM-as-judge provides scalable assessment, it may introduce biases. Our human evaluation validates its reliability, showing high consistency with human judgment in model ranking. We conduct two studies: (1) 3 annotators evaluate 900 outputs from DeepSeek-V3.2, and (2) 100 randomly sampled outputs each from GLM-4.6, Gemini-2.5-Pro, LongCat-Flash-Thinking, and Qwen3-235B-A22B. Human scores demonstrate substantial inter-rater reliability (Cohen's $\kappa$ = 0.79, 95\% CI: $[0.75, 0.83]$, 87.91\% raw agreement). 

\subsection{Ablation Study}
We conduct an ablation study
on tool integration---a core design element for \emph{agentic search} in local life services. Specifically, we evaluate three configurations across top-performing \emph{LRM}s (\ie, Gemini-2.5-Pro, GLM-4.6, DeepSeek-V3.2): (1) \emph{LRM}-only (no tools), (2) \emph{LRM} + \rag (merchant retrieval only), and (3) \emph{LRM} + \rag + \emph{web search} (full toolchain). The goal is to quantify how each tool impacts answer and trajectory quality.
As shown in Table~\ref{tab:model_performance}, incorporating web search improves answer quality but reduces faithfulness. On average, it boosts Correctness by 8.92 pp and Completeness by 7.41 pp, while decreasing Faithfulness by 2.83 pp. This trade-off is most evident in Gemini-2.5-Pro, which gains 9.23 pp in Correctness but loses 3.72 pp in Faithfulness. These results highlight web search's dual role in enabling critical information retrieval and introducing noise in multi-hop reasoning.

\begin{figure}[t]
    \centering
    \includegraphics[width=0.92\linewidth]{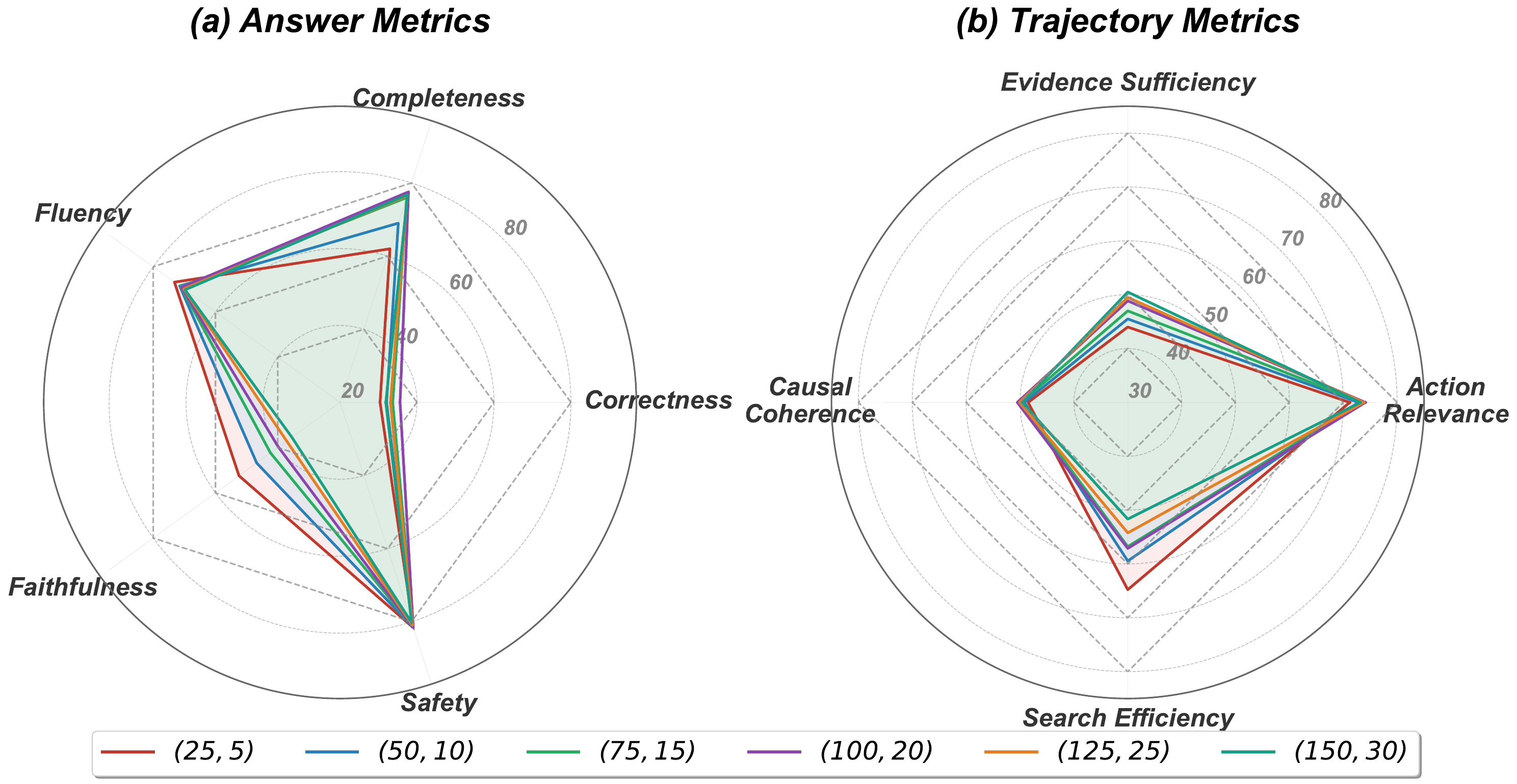}
    \caption{Impact of Retrieval Parameters (top-\textit{N}, top-\textit{K}) of \rag on Evaluation Metrics.}
    \label{fig:retrieval_sensitivity}
\end{figure}

\subsection{Sensitivity Study}
To investigate the impact of maximum conversation rounds $N$ on tool usage and model performance, we conduct a sensitivity study using DeepSeek-V3.2, the best-performing model, with varying $N$ from 2 to 8. 
As shown in Table~\ref{tab:sensitivity}, the optimal configuration is $N=5$, achieving the highest Correctness (35.60\%). When $N < 5$, insufficient rounds limit information gathering, resulting in lower correctness (30.56\%--34.22\%). When $N > 5$, despite increased tool usage, correctness drops (34.89\%--32.44\%), suggesting excessive rounds introduce noise that hinders answer quality. In terms of computational cost, tool calls and conversation length increase with $N$, peaking at $N=8$ (4.02 tool calls, 5.01 rounds). Therefore, $N=5$ represents the optimal balance of performance and efficiency.

We analyze how retrieval hyperparameters affect downstream agentic search performance by systematically varying \rag's two key parameters: top-\emph{N} (initial candidates from dense retrieval) and top-\emph{K} (merchants returned after reranking). We evaluate 6 paired configurations: (top-\emph{N}, top-\emph{K}) $\in$ \{(25,5), (50,10), (75,15), (100,20), (125,25), (150,30)\} using DeepSeek-V3.2 on 900 multi-hop QA with fixed settings (maximum rounds \emph{N}=5, temperature=0). As shown in Figure~\ref{fig:retrieval_sensitivity}, the optimal configuration (top-\emph{N}=100, top-\emph{K}=20) achieves Correctness of 35.6\% (ANOVA: $F=259.73, p<0.001$); however, increasing to (150,30) degrades Correctness by 3.7\%, Faithfulness by 4.3\%, and Search Efficiency by 5.4\%, indicating information overload effects.

\subsection{In-Depth Analysis}

\begin{figure}[t]
    \centering
    \includegraphics[width=0.95\linewidth]{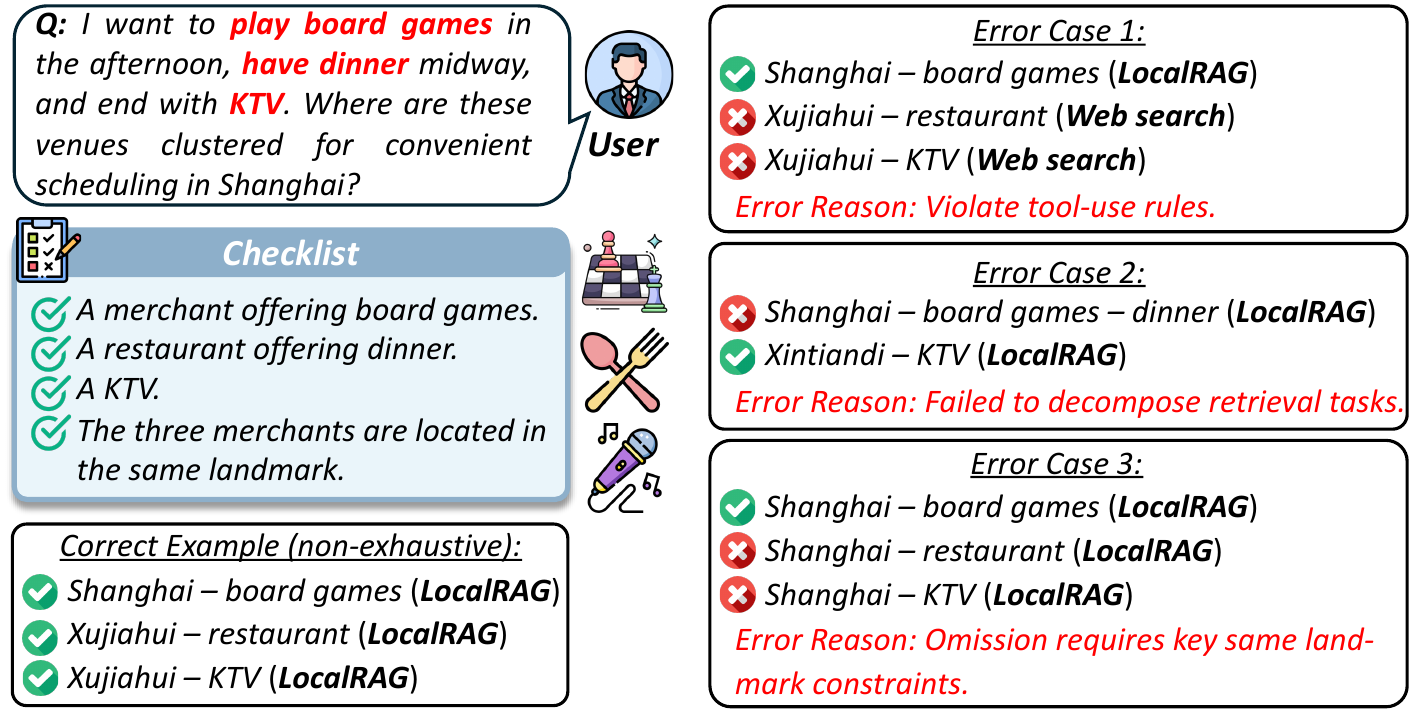}
    \caption{Failure Case Studies. A typical chain event: the 2\textsuperscript{nd} and 3\textsuperscript{rd} hops depend on the 1\textsuperscript{st} hop.}
    \label{fig:case-study}
\end{figure}
We study factors affecting \emph{LRM} performance on agentic search tasks in local life services. Failure analysis is conducted on 580 DeepSeek-V3.2 failure trajectories 
from a single experimental run, annotated by 3 domain experts. Inter-annotator agreement achieves Cohen's $\kappa = 0.84$. Our main findings are:

\textbf{(1) Unstable tool call strategy (38.6\%).} Lack of a unified strategy for tool call timing and frequency leads to insufficient tool calls or wasted budget. \emph{LRM}s may also violate tool-use rules, such as using web search for merchant queries or making redundant tool calls within a single dialogue turn.
\textbf{(2) Missing explicit multi-hop planning (30.9\%).} \emph{LRM}s lacking hop sequences fail to decompose tasks for retrieval or tool use. Without explicit hop targets, sub-goals remain unordered, causing empty matches and higher coverage failures in multi-hop tasks. Retrieval failures cascade through steps due to the absence of validation or backtracking.
\textbf{(3) Query generation dilutes critical constraints (11.7\%).}  Naively generated long-form queries often omit key constraints, weakening retrieval signals and reducing recall of relevant candidates.
\textbf{(4) Long-context noise (18.8\%).} Lengthy search results and information sent to \emph{LRM}s bring irrelevant data, reduce the ability to distinguish relevant results, and can exceed context window limits.

\section{Conclusion}




We propose \textbf{\bench}, a benchmark for \emph{agentic search} in local life services evaluating multi-hop reasoning. It contains 1,354,185 entries and 900 multi-hop QA tasks across 9 Chinese cities and 6 service categories. Experiments on 16 leading LLMs show current models struggle, with the best model DeepSeek-V3.2 achieving only 35.60\% correctness. We open-source \bench and \eval to advance domain-specific agentic search.


\bibliographystyle{ACM-Reference-Format}
\bibliography{citation}

@misc{gou2025mind2web2evaluatingagentic,
      title={Mind2Web 2: Evaluating Agentic Search with Agent-as-a-Judge}, 
      author={Boyu Gou and Zanming Huang and Yuting Ning and Yu Gu and Michael Lin and Weijian Qi and Andrei Kopanev and Botao Yu and Bernal Jiménez Gutiérrez and Yiheng Shu and Chan Hee Song and Jiaman Wu and Shijie Chen and Hanane Nour Moussa and Tianshu Zhang and Jian Xie and Yifei Li and Tianci Xue and Zeyi Liao and Kai Zhang and Boyuan Zheng and Zhaowei Cai and Viktor Rozgic and Morteza Ziyadi and Huan Sun and Yu Su},
      year={2025},
      eprint={2506.21506},
      archivePrefix={arXiv},
      primaryClass={cs.AI},
      url={https://arxiv.org/abs/2506.21506}, 
}

@article{jaech2024openai,
  title={Openai o1 system card},
  author={Jaech, Aaron and Kalai, Adam and Lerer, Adam and Richardson, Adam and El-Kishky, Ahmed and Low, Aiden and Helyar, Alec and Madry, Aleksander and Beutel, Alex and Carney, Alex and others},
  journal={arXiv preprint arXiv:2412.16720},
  year={2024}
}

@article{jin2025search,
  title={Search-r1: Training llms to reason and leverage search engines with reinforcement learning},
  author={Jin, Bowen and Zeng, Hansi and Yue, Zhenrui and Yoon, Jinsung and Arik, Sercan and Wang, Dong and Zamani, Hamed and Han, Jiawei},
  journal={arXiv preprint arXiv:2503.09516},
  year={2025}
}

@article{song2025r1,
  title={R1-searcher: Incentivizing the search capability in llms via reinforcement learning},
  author={Song, Huatong and Jiang, Jinhao and Min, Yingqian and Chen, Jie and Chen, Zhipeng and Zhao, Wayne Xin and Fang, Lei and Wen, Ji-Rong},
  journal={arXiv preprint arXiv:2503.05592},
  year={2025}
}

@article{li2025search,
  title={Search-o1: Agentic search-enhanced large reasoning models},
  author={Li, Xiaoxi and Dong, Guanting and Jin, Jiajie and Zhang, Yuyao and Zhou, Yujia and Zhu, Yutao and Zhang, Peitian and Dou, Zhicheng},
  journal={arXiv preprint arXiv:2501.05366},
  year={2025}
}

@article{li2025webthinker,
  title={Webthinker: Empowering large reasoning models with deep research capability},
  author={Li, Xiaoxi and Jin, Jiajie and Dong, Guanting and Qian, Hongjin and Zhu, Yutao and Wu, Yongkang and Wen, Ji-Rong and Dou, Zhicheng},
  journal={arXiv preprint arXiv:2504.21776},
  year={2025}
}

@article{li2025towards,
  title={Towards AI Search Paradigm},
  author={Li, Yuchen and Cai, Hengyi and Kong, Rui and Chen, Xinran and Chen, Jiamin and Yang, Jun and Zhang, Haojie and Li, Jiayi and Wu, Jiayi and Chen, Yiqun and others},
  journal={arXiv preprint arXiv:2506.17188},
  year={2025}
}

@article{yang2025agentic,
  title={Agentic web: Weaving the next web with ai agents},
  author={Yang, Yingxuan and Ma, Mulei and Huang, Yuxuan and Chai, Huacan and Gong, Chenyu and Geng, Haoran and Zhou, Yuanjian and Wen, Ying and Fang, Meng and Chen, Muhao and others},
  journal={arXiv preprint arXiv:2507.21206},
  year={2025}
}

@misc{tongyidr,
  author={Tongyi DeepResearch Team},
  title={Tongyi-DeepResearch},
  year={2025},
  howpublished={\url{https://github.com/Alibaba-NLP/DeepResearch}}
}

@inproceedings{lan2025benchmarking,
  title={Benchmarking and advancing large language models for local life services},
  author={Lan, Xiaochong and Feng, Jie and Lei, Jiahuan and Shi, Xinlei and Li, Yong},
  booktitle={Proceedings of the 31st ACM SIGKDD Conference on Knowledge Discovery and Data Mining V. 2},
  pages={4566--4577},
  year={2025}
}

@inproceedings{10.1145/3580305.3599874,
author = {Lan, Xiaochong and Gao, Chen and Wen, Shiqi and Chen, Xiuqi and Che, Yingge and Zhang, Han and Wei, Huazhou and Luo, Hengliang and Li, Yong},
title = {NEON: Living Needs Prediction System in Meituan},
year = {2023},
isbn = {9798400701030},
publisher = {Association for Computing Machinery},
address = {New York, NY, USA},
url = {https://doi.org/10.1145/3580305.3599874},
doi = {10.1145/3580305.3599874},
booktitle = {Proceedings of the 29th ACM SIGKDD Conference on Knowledge Discovery and Data Mining},
pages = {5292--5302},
numpages = {11},
location = {Long Beach, CA, USA},
series = {KDD '23}
}

@inproceedings{10.1145/3534678.3539122,
author = {Li, Yinfeng and Gao, Chen and Du, Xiaoyi and Wei, Huazhou and Luo, Hengliang and Jin, Depeng and Li, Yong},
title = {Automatically Discovering User Consumption Intents in Meituan},
year = {2022},
isbn = {9781450393850},
publisher = {Association for Computing Machinery},
address = {New York, NY, USA},
url = {https://doi.org/10.1145/3534678.3539122},
doi = {10.1145/3534678.3539122},
booktitle = {Proceedings of the 28th ACM SIGKDD Conference on Knowledge Discovery and Data Mining},
pages = {3259--3269},
numpages = {11},
location = {Washington DC, USA},
series = {KDD '22}
}

@inproceedings{kusk2025flexible,
  title={Flexible Platforms? An Ethnographic Study of Flexible Scheduling in Platform-Mediated Delivery},
  author={Kusk, Kalle},
  booktitle={Proceedings of the 2025 CHI Conference on Human Factors in Computing Systems},
  pages={1--11},
  year={2025}
}

@inproceedings{liu2025mrgrp,
  title={Mrgrp: Empowering courier route prediction in food delivery service with multi-relational graph},
  author={Liu, Chang and Yan, Huan and Sui, Hongjie and Wen, Haomin and Yuan, Yuan and Han, Yuyang and Liao, Hongsen and Ding, Xuetao and Hao, Jinghua and Li, Yong},
  booktitle={Companion Proceedings of the ACM on Web Conference 2025},
  pages={364--373},
  year={2025}
}

@misc{sun2025simpledeepsearch,
      title={SimpleDeepSearcher: Deep Information Seeking via Web-Powered Reasoning Trajectory Synthesis}, 
      author={Shuang Sun and Huatong Song and Yuhao Wang and Ruiyang Ren and Jinhao Jiang and Junjie Zhang and Fei Bai and Jia Deng and Wayne Xin Zhao and Zheng Liu and Lei Fang and Zhongyuan Wang and Ji-Rong Wen},
      year={2025},
      eprint={2505.16834},
      archivePrefix={arXiv},
      primaryClass={cs.CL},
      url={https://arxiv.org/abs/2505.16834}, 
}

@misc{zhang2025evolvesearch,
      title={EvolveSearch: An Iterative Self-Evolving Search Agent}, 
      author={Dingchu Zhang and Yida Zhao and Jialong Wu and Baixuan Li and Wenbiao Yin and Liwen Zhang and Yong Jiang and Yufeng Li and Kewei Tu and Pengjun Xie and Fei Huang},
      year={2025},
      eprint={2505.22501},
      archivePrefix={arXiv},
      primaryClass={cs.CL},
      url={https://arxiv.org/abs/2505.22501}, 
}

@misc{shi2025pangudeepdiveradaptivesearch,
      title={DeepDiver: Adaptive Search Intensity Scaling via Open-Web Reinforcement Learning},
      author={Wenxuan Shi and Haochen Tan and Chuqiao Kuang and Xiaoguang Li and Xiaozhe Ren and Chen Zhang and Hanting Chen and Yasheng Wang and Lifeng Shang and Fisher Yu and Yunhe Wang},
  year={2025},
  eprint={2505.24332},
  archivePrefix={arXiv},
  primaryClass={cs.CL},
  url={https://arxiv.org/abs/2505.24332}
}

@misc{shi2025toollearningwildempowering,
      title={Tool Learning in the Wild: Empowering Language Models as Automatic Tool Agents}, 
      author={Zhengliang Shi and Shen Gao and Lingyong Yan and Yue Feng and Xiuyi Chen and Zhumin Chen and Dawei Yin and Suzan Verberne and Zhaochun Ren},
      year={2025},
      eprint={2405.16533},
      archivePrefix={arXiv},
      primaryClass={cs.CL},
      url={https://arxiv.org/abs/2405.16533}, 
}

@misc{wei2025browsecomps,
      title={BrowseComp: A Simple Yet Challenging Benchmark for Browsing Agents}, 
      author={Jason Wei and Zhiqing Sun and Spencer Papay and Scott McKinney and Jeffrey Han and Isa Fulford and Hyung Won Chung and Alex Tachard Passos and William Fedus and Amelia Glaese},
      year={2025},
      eprint={2504.12516},
      archivePrefix={arXiv},
      primaryClass={cs.CL},
      url={https://arxiv.org/abs/2504.12516}, 
}

@misc{zhou2025browsecompzhbenchmarking,
      title={BrowseComp-ZH: Benchmarking Web Browsing Ability of Large Language Models in Chinese}, 
      author={Peilin Zhou and Bruce Leon and Xiang Ying and Can Zhang and Yifan Shao and Qichen Ye and Dading Chong and Zhiling Jin and Chenxuan Xie and Meng Cao and Yuxin Gu and Sixin Hong and Jing Ren and Jian Chen and Chao Liu and Yining Hua},
      year={2025},
      eprint={2504.19314},
      archivePrefix={arXiv},
      primaryClass={cs.CL},
      url={https://arxiv.org/abs/2504.19314}, 
}

@article{zhang2025web,
  title={From Web Search towards Agentic Deep Research: Incentivizing Search with Reasoning Agents},
  author={Zhang, Weizhi and Li, Yangning and Bei, Yuanchen and Luo, Junyu and Wan, Guancheng and Yang, Liangwei and Xie, Chenxuan and Yang, Yuyao and Huang, Wei-Chieh and Miao, Chunyu and others},
  journal={arXiv preprint arXiv:2506.18959},
  year={2025}
}

@inproceedings{mo2025conversational,
  title={Conversational search: From fundamentals to frontiers in the LLM era},
  author={Mo, Fengran and Meng, Chuan and Aliannejadi, Mohammad and Nie, Jian-Yun},
  booktitle={Proceedings of the 48th International ACM SIGIR Conference on Research and Development in Information Retrieval},
  pages={4094--4097},
  year={2025}
}

@misc{xu2025towards,
      title={Towards Large Reasoning Models: A Survey of Reinforced Reasoning with Large Language Models}, 
      author={Fengli Xu and Qianyue Hao and Zefang Zong and Jingwei Wang and Yunke Zhang and Jingyi Wang and Xiaochong Lan and Jiahui Gong and Tianjian Ouyang and Fanjin Meng and Chenyang Shao and Yuwei Yan and Qinglong Yang and Yiwen Song and Sijian Ren and Xinyuan Hu and Yu Li and Jie Feng and Chen Gao and Yong Li},
      year={2025},
      eprint={2501.09686},
      archivePrefix={arXiv},
      primaryClass={cs.AI},
      url={https://arxiv.org/abs/2501.09686}, 
}

@misc{ferrag2025llm,
      title={From LLM Reasoning to Autonomous AI Agents: A Comprehensive Review}, 
      author={Mohamed Amine Ferrag and Norbert Tihanyi and Merouane Debbah},
      year={2025},
      eprint={2504.19678},
      archivePrefix={arXiv},
      primaryClass={cs.AI},
      url={https://arxiv.org/abs/2504.19678}, 
}

@misc{wu2025agentic,
      title={Agentic Reasoning: A Streamlined Framework for Enhancing LLM Reasoning with Agentic Tools}, 
      author={Junde Wu and Jiayuan Zhu and Yuyuan Liu and Min Xu and Yueming Jin},
      year={2025},
      eprint={2502.04644},
      archivePrefix={arXiv},
      primaryClass={cs.AI},
      url={https://arxiv.org/abs/2502.04644}, 
}

@article{ke2025survey,
  title={A survey of frontiers in llm reasoning: Inference scaling, learning to reason, and agentic systems},
  author={Ke, Zixuan and Jiao, Fangkai and Ming, Yifei and Nguyen, Xuan-Phi and Xu, Austin and Long, Do Xuan and Li, Minzhi and Qin, Chengwei and Wang, Peifeng and Savarese, Silvio and others},
  journal={arXiv preprint arXiv:2504.09037},
  year={2025}
}

@misc{zhou2024webarenarealistic,
      title={WebArena: A Realistic Web Environment for Building Autonomous Agents}, 
      author={Shuyan Zhou and Frank F. Xu and Hao Zhu and Xuhui Zhou and Robert Lo and Abishek Sridhar and Xianyi Cheng and Tianyue Ou and Yonatan Bisk and Daniel Fried and Uri Alon and Graham Neubig},
      year={2024},
      eprint={2307.13854},
      archivePrefix={arXiv},
      primaryClass={cs.AI},
      url={https://arxiv.org/abs/2307.13854}, 
}

@misc{yao2023webshopscalable,
      title={WebShop: Towards Scalable Real-World Web Interaction with Grounded Language Agents}, 
      author={Shunyu Yao and Howard Chen and John Yang and Karthik Narasimhan},
      year={2023},
      eprint={2207.01206},
      archivePrefix={arXiv},
      primaryClass={cs.CL},
      url={https://arxiv.org/abs/2207.01206}, 
}

@article{yuan2025videodeepresearch,
  title={VideoExplorer: Think With Videos For Agentic Long-Video Understanding},
  author={Yuan, Huaying and Liu, Zheng and Zhou, Junjie and Qian, Hongjin and Shu, Yan and Sebe, Nicu and Wen, Ji-Rong and Dou, Zhicheng},
  journal={arXiv preprint arXiv:2506.10821},
  year={2025}
}

@article{du2025deepresearch,
  title={DeepResearch Bench: A Comprehensive Benchmark for Deep Research Agents},
  author={Du, Mingxuan and Xu, Benfeng and Zhu, Chiwei and Wang, Xiaorui and Mao, Zhendong},
  journal={arXiv preprint arXiv:2506.11763},
  year={2025}
}

@article{tan2024proxyqa,
  title={Proxyqa: An alternative framework for evaluating long-form text generation with large language models},
  author={Tan, Haochen and Guo, Zhijiang and Shi, Zhan and Xu, Lu and Liu, Zhili and Feng, Yunlong and Li, Xiaoguang and Wang, Yasheng and Shang, Lifeng and Liu, Qun and others},
  journal={arXiv preprint arXiv:2401.15042},
  year={2024}
}

@article{rosset2024researchy,
  title={Researchy questions: A dataset of multi-perspective, decompositional questions for llm web agents},
  author={Rosset, Corby and Chung, Ho-Lam and Qin, Guanghui and Chau, Ethan C and Feng, Zhuo and Awadallah, Ahmed and Neville, Jennifer and Rao, Nikhil},
  journal={arXiv preprint arXiv:2402.17896},
  year={2024}
}

@article{wei2024long,
  title={Long-form factuality in large language models},
  author={Wei, Jerry and Yang, Chengrun and Song, Xinying and Lu, Yifeng and Hu, Nathan and Huang, Jie and Tran, Dustin and Peng, Daiyi and Liu, Ruibo and Huang, Da and others},
  journal={Advances in Neural Information Processing Systems},
  volume={37},
  pages={80756--80827},
  year={2024}
}

@article{wadden2020fact,
  title={Fact or fiction: Verifying scientific claims},
  author={Wadden, David and Lin, Shanchuan and Lo, Kyle and Wang, Lucy Lu and van Zuylen, Madeleine and Cohan, Arman and Hajishirzi, Hannaneh},
  journal={arXiv preprint arXiv:2004.14974},
  year={2020}
}

@article{jiang2020hover,
  title={HoVer: A dataset for many-hop fact extraction and claim verification},
  author={Jiang, Yichen and Bordia, Shikha and Zhong, Zheng and Dognin, Charles and Singh, Maneesh and Bansal, Mohit},
  journal={arXiv preprint arXiv:2011.03088},
  year={2020}
}

@article{wang2024mfc,
  title={Mfc-bench: Benchmarking multimodal fact-checking with large vision-language models},
  author={Wang, Shengkang and Lin, Hongzhan and Luo, Ziyang and Ye, Zhen and Chen, Guang and Ma, Jing},
  journal={arXiv preprint arXiv:2406.11288},
  year={2024}
}

@article{bosse2025deep,
  title={Deep Research Bench: Evaluating AI Web Research Agents},
  author={Bosse, Nikos I and Evans, Jon and Gambee, Robert G and Hnyk, Daniel and M{\"u}hlbacher, Peter and Phillips, Lawrence and Schwarz, Dan and Wildman, Jack and others},
  journal={arXiv preprint arXiv:2506.06287},
  year={2025}
}

@article{xi2025infodeepseek,
  title={InfoDeepSeek: Benchmarking Agentic Information Seeking for Retrieval-Augmented Generation},
  author={Xi, Yunjia and Lin, Jianghao and Zhu, Menghui and Xiao, Yongzhao and Ou, Zhuoying and Liu, Jiaqi and Wan, Tong and Chen, Bo and Liu, Weiwen and Wang, Yasheng and others},
  journal={arXiv preprint arXiv:2505.15872},
  year={2025}
}

@article{pham2025sealqa,
  title={SealQA: Raising the Bar for Reasoning in Search-Augmented Language Models},
  author={Pham, Thinh and Nguyen, Nguyen and Zunjare, Pratibha and Chen, Weiyuan and Tseng, Yu-Min and Vu, Tu},
  journal={arXiv preprint arXiv:2506.01062},
  year={2025}
}

@article{yang2018hotpotqa,
  title={HotpotQA: A dataset for diverse, explainable multi-hop question answering},
  author={Yang, Zhilin and Qi, Peng and Zhang, Saizheng and Bengio, Yoshua and Cohen, William W and Salakhutdinov, Ruslan and Manning, Christopher D},
  journal={arXiv preprint arXiv:1809.09600},
  year={2018}
}

@article{ho2020constructing,
  title={Constructing a multi-hop qa dataset for comprehensive evaluation of reasoning steps},
  author={Ho, Xanh and Nguyen, Anh-Khoa Duong and Sugawara, Saku and Aizawa, Akiko},
  journal={arXiv preprint arXiv:2011.01060},
  year={2020}
}

@article{press2022measuring,
  title={Measuring and narrowing the compositionality gap in language models},
  author={Press, Ofir and Zhang, Muru and Min, Sewon and Schmidt, Ludwig and Smith, Noah A and Lewis, Mike},
  journal={arXiv preprint arXiv:2210.03350},
  year={2022}
}

@article{trivedi2022musique,
  title={MuSiQue: Multihop Questions via Single-hop Question Composition},
  author={Trivedi, Harsh and Balasubramanian, Niranjan and Khot, Tushar and Sabharwal, Ashish},
  journal={Transactions of the Association for Computational Linguistics},
  volume={10},
  pages={539--554},
  year={2022},
  publisher={MIT Press One Broadway, 12th Floor, Cambridge, Massachusetts 02142, USA~…}
}

@misc{anthropic2025claude41,
  author={{Anthropic}},
  title={Introducing Claude Opus 4.5},
  year={2025},
  url={https://www.anthropic.com/news/claude-opus-4-5}
}

@misc{deepseek2025v3,
  author={{DeepSeek-AI}},
  title={DeepSeek-V3.2 Release},
  year={2025},
  url={https://api-docs.deepseek.com/news/news251201}
}

@article{team2025longcat,
  title={LongCat-Flash Technical Report},
  author={Team, Meituan LongCat and Li, Bei and Lei, Bingye and Wang, Bo and Rong, Bolin and Wang, Chao and Zhang, Chao and Gao, Chen and Zhang, Chen and Sun, Cheng and others},
  journal={arXiv preprint arXiv:2509.01322},
  year={2025}
}

@article{zeng2025glm,
  title={Glm-4.5: Agentic, reasoning, and coding (arc) foundation models},
  author={Zeng, Aohan and Lv, Xin and Zheng, Qinkai and Hou, Zhenyu and Chen, Bin and Xie, Chengxing and Wang, Cunxiang and Yin, Da and Zeng, Hao and Zhang, Jiajie and others},
  journal={arXiv preprint arXiv:2508.06471},
  year={2025}
}

@article{comanici2025gemini,
  title={Gemini 2.5: Pushing the frontier with advanced reasoning, multimodality, long context, and next generation agentic capabilities},
  author={Comanici, Gheorghe and Bieber, Eric and Schaekermann, Mike and Pasupat, Ice and Sachdeva, Noveen and Dhillon, Inderjit and Blistein, Marcel and Ram, Ori and Zhang, Dan and Rosen, Evan and others},
  journal={arXiv preprint arXiv:2507.06261},
  year={2025}
}

@misc{openai2025gpt4-1,
    author = {{OpenAI}},
    title = {GPT-4.1 Model},
    year = {2025},
    url = {https://openai.com/index/gpt-4-1/},
}

@article{yang2025qwen3,
  title={Qwen3 technical report},
  author={Yang, An and Li, Anfeng and Yang, Baosong and Zhang, Beichen and Hui, Binyuan and Zheng, Bo and Yu, Bowen and Gao, Chang and Huang, Chengen and Lv, Chenxu and others},
  journal={arXiv preprint arXiv:2505.09388},
  year={2025}
}

@inproceedings{liu2021pre,
  title={Pre-trained language model for web-scale retrieval in baidu search},
  author={Liu, Yiding and Lu, Weixue and Cheng, Suqi and Shi, Daiting and Wang, Shuaiqiang and Cheng, Zhicong and Yin, Dawei},
  booktitle={Proceedings of the 27th ACM SIGKDD Conference on Knowledge Discovery \& Data Mining},
  pages={3365--3375},
  year={2021}
}

@article{zhang2025qwen3,
  title={Qwen3 Embedding: Advancing Text Embedding and Reranking Through Foundation Models},
  author={Zhang, Yanzhao and Li, Mingxin and Long, Dingkun and Zhang, Xin and Lin, Huan and Yang, Baosong and Xie, Pengjun and Yang, An and Liu, Dayiheng and Lin, Junyang and others},
  journal={arXiv preprint arXiv:2506.05176},
  year={2025}
}

@inproceedings{liu2025queries,
  title={Queries Are Not Alone: Clustering Text Embeddings for Video Search},
  author={Liu, Peiyang and Wang, Xi and Cui, Ziqiang and Ye, Wei},
  booktitle={Proceedings of the 48th International ACM SIGIR Conference on Research and Development in Information Retrieval},
  pages={874--883},
  year={2025}
}

@misc{chai2025rlfactory,
      title={RLFactory: A Plug-and-Play Reinforcement Learning Post-Training Framework for LLM Multi-Turn Tool-Use}, 
      author={Jiajun Chai and Guojun Yin and Zekun Xu and Chuhuai Yue and Yi Jia and Siyu Xia and Xiaohan Wang and Jiwen Jiang and Xiaoguang Li and Chengqi Dong and Hang He and Wei Lin},
      year={2025},
      eprint={2509.06980},
      archivePrefix={arXiv},
      primaryClass={cs.LG},
      url={https://arxiv.org/abs/2509.06980}, 
}

@misc{yue2025promoting,
      title={Promoting Efficient Reasoning with Verifiable Stepwise Reward}, 
      author={Chuhuai Yue and Chengqi Dong and Yinan Gao and Hang He and Jiajun Chai and Guojun Yin and Wei Lin},
      year={2025},
      eprint={2508.10293},
      archivePrefix={arXiv},
      primaryClass={cs.AI},
      url={https://arxiv.org/abs/2508.10293}, 
}

@inproceedings{yue2025uiorchestra,
  title={{UIO}rchestra: Generating High-Fidelity Code from {UI} Designs with a Multi-agent System},
  author={Yue, Chuhuai and Chai, Jiajun and Zhang, Yufei and Ding, Zixiang and Liang, Xihao and Wang, Peixin and Chen, Shihai and Yixuan, Wang and Wangyanping and Yin, Guojun and Lin, Wei},
  booktitle={Findings of the Association for Computational Linguistics: EMNLP 2025},
  pages={2769--2782},
  year={2025},
  publisher={Association for Computational Linguistics},
  url={https://aclanthology.org/2025.findings-emnlp.150/}
}

@misc{chen2025toolforgedatasynthesispipeline,
      title={ToolForge: A Data Synthesis Pipeline for Multi-Hop Search without Real-World APIs}, 
      author={Hao Chen and Zhexin Hu and Jiajun Chai and Haocheng Yang and Hang He and Xiaohan Wang and Wei Lin and Luhang Wang and Guojun Yin and Zhuofeng zhao},
      year={2025},
      eprint={2512.16149},
      archivePrefix={arXiv},
      primaryClass={cs.AI},
      url={https://arxiv.org/abs/2512.16149}, 
}

@misc{dong2025trainingmultiimagevisionagents,
      title={Training Multi-Image Vision Agents via End2End Reinforcement Learning}, 
      author={Chengqi Dong and Chuhuai Yue and Hang He and Rongge Mao and Fenghe Tang and S Kevin Zhou and Zekun Xu and Xiaohan Wang and Jiajun Chai and Wei Lin and Guojun Yin},
      year={2025},
      eprint={2512.08980},
      archivePrefix={arXiv},
      primaryClass={cs.CV},
      url={https://arxiv.org/abs/2512.08980}, 
}

@misc{lan2025deepwidesearch,
      title={DeepWideSearch: Benchmarking Depth and Width in Agentic Information Seeking},
      author={Tian Lan and Bin Zhu and Qianghuai Jia and Junyang Ren and Haijun Li and Longyue Wang and Zhao Xu and Weihua Luo and Kaifu Zhang},
      year={2025},
      eprint={2510.20168},
      archivePrefix={arXiv},
      primaryClass={cs.CL},
      url={https://arxiv.org/abs/2510.20168},
}

@article{cohen1960coefficient,
  title={A coefficient of agreement for nominal scales},
  author={Cohen, Jacob},
  journal={Educational and psychological measurement},
  volume={20},
  number={1},
  pages={37--46},
  year={1960}
}

@misc{wang2025lmars,
      title={L-MARS: Legal Multi-Agent Workflow with Orchestrated Reasoning and Agentic Search},
      author={Ziqi Wang and Boqin Yuan},
      year={2025},
      eprint={2509.00761},
      archivePrefix={arXiv},
      primaryClass={cs.AI},
      url={https://arxiv.org/abs/2509.00761},
}

\begin{appendices}
\newpage
\section*{APPENDIX}
\section{Dataset Construction and Task Design}
\label{appendix:dataset}
\subsection{\bench Task Details}

\begin{figure}[htbp]
    \centering
    \resizebox{0.48\textwidth}{!}{%
    \begin{minipage}{\textwidth}
        \centering
        \begin{subfigure}[b]{0.49\columnwidth}
            \centering
            \includegraphics[width=0.9\textwidth]{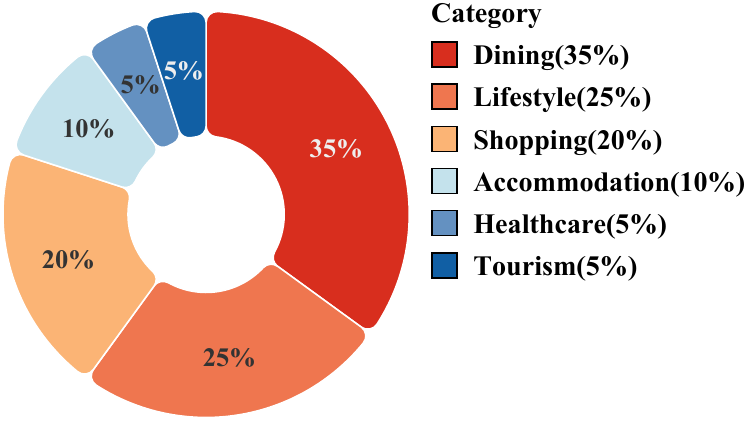}
            \caption{Standardized category distribution of real-world data}
            \label{fig:proportion}
        \end{subfigure}\hfill
        \begin{subfigure}[b]{0.49\columnwidth}
            \centering
            \includegraphics[width=0.9\textwidth]{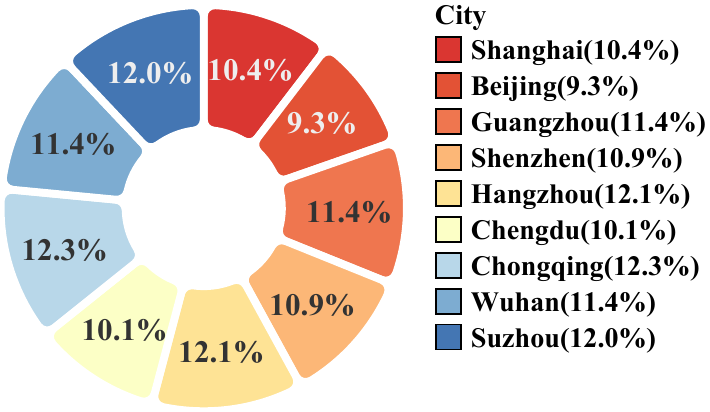}
            \caption{City distribution}
            \label{fig:city_distribution}
        \end{subfigure}
        
        \vspace{1em}
        
        \begin{subfigure}[b]{0.27\columnwidth}
            \centering
            \includegraphics[width=0.9\textwidth]{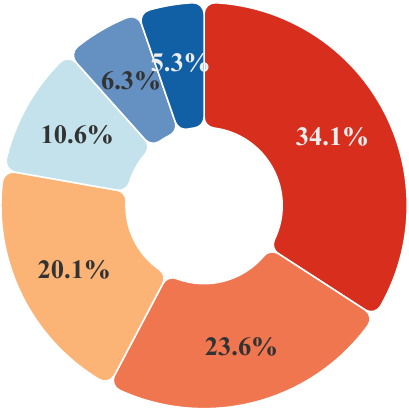}
            \caption{Shanghai}
            \label{fig:shanghai}
        \end{subfigure}\hfill
        \begin{subfigure}[b]{0.27\columnwidth}
            \centering
            \includegraphics[width=0.9\textwidth]{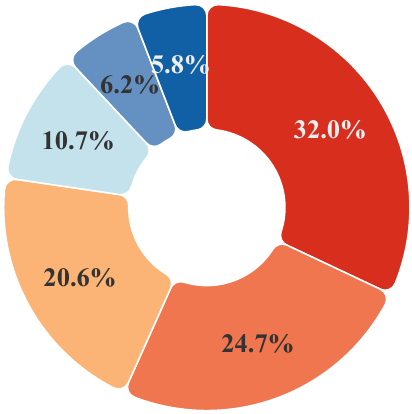}
            \caption{Beijing}
            \label{fig:beijing}
        \end{subfigure}\hfill
        \begin{subfigure}[b]{0.27\columnwidth}
            \centering
            \includegraphics[width=0.9\textwidth]{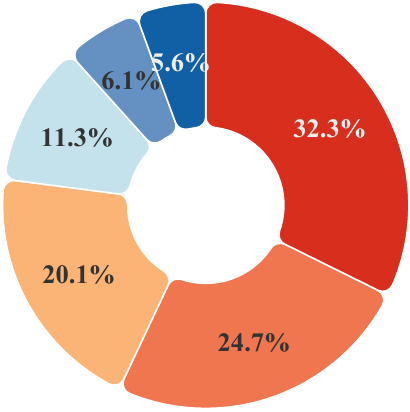}
            \caption{Guangzhou}
            \label{fig:guangzhou}
        \end{subfigure}
        \begin{subfigure}[b]{0.18\columnwidth}
            \centering
            \includegraphics[width=0.9\textwidth]{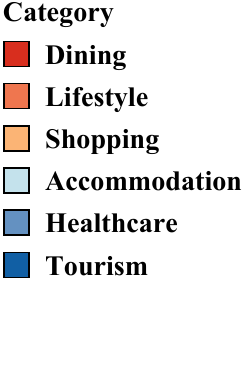}
        \end{subfigure}
        
        \vspace{0.5em}
        
        \begin{subfigure}[b]{0.27\columnwidth}
            \centering
            \includegraphics[width=0.9\textwidth]{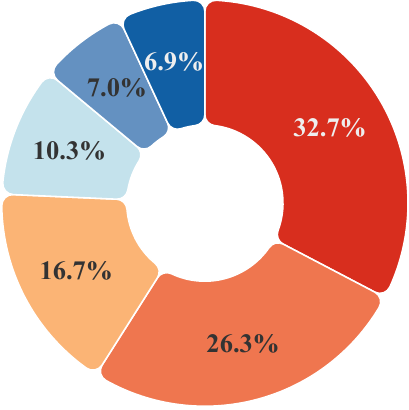}
            \caption{Shenzhen}
            \label{fig:shenzhen}
        \end{subfigure}\hfill
        \begin{subfigure}[b]{0.27\columnwidth}
            \centering
            \includegraphics[width=0.9\textwidth]{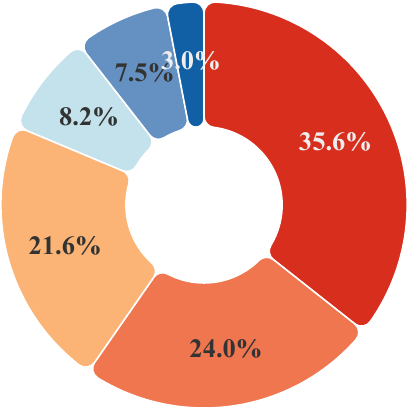}
            \caption{Hangzhou}
            \label{fig:hangzhou}
        \end{subfigure}\hfill
        \begin{subfigure}[b]{0.27\columnwidth}
            \centering
            \includegraphics[width=0.9\textwidth]{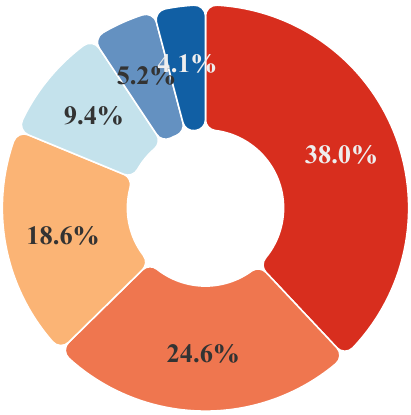}
            \caption{Chengdu}
            \label{fig:chengdu}
        \end{subfigure}
        \begin{subfigure}[b]{0.18\columnwidth}
            \centering
            \includegraphics[width=0.9\textwidth]{figure/tuanshi-1.pdf}
        \end{subfigure}
        
        \vspace{0.5em}
        
        \begin{subfigure}[b]{0.27\columnwidth}
            \centering
            \includegraphics[width=0.9\textwidth]{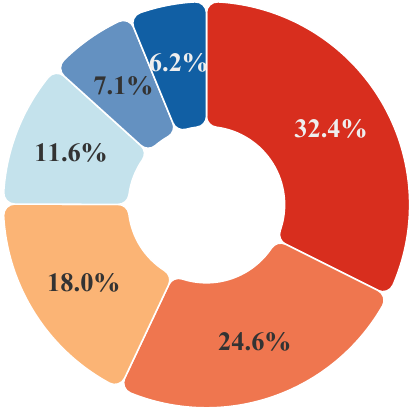}
            \caption{Chongqing}
            \label{fig:chongqing}
        \end{subfigure}\hfill
        \begin{subfigure}[b]{0.27\columnwidth}
            \centering
            \includegraphics[width=0.9\textwidth]{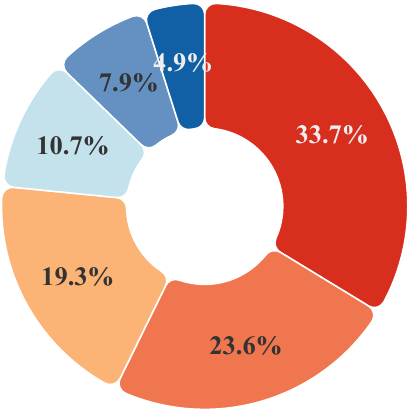}
            \caption{Wuhan}
            \label{fig:wuhan}
        \end{subfigure}\hfill
        \begin{subfigure}[b]{0.27\columnwidth}
            \centering
            \includegraphics[width=0.9\textwidth]{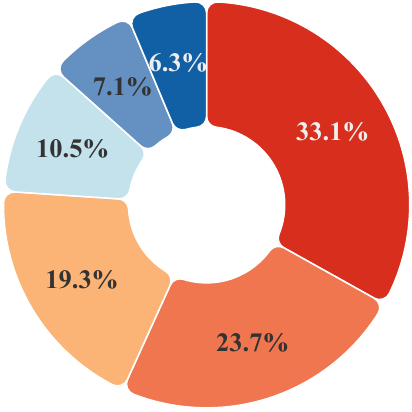}
            \caption{Suzhou}
            \label{fig:suzhou}
        \end{subfigure}
        \begin{subfigure}[b]{0.18\columnwidth}
            \centering
            \includegraphics[width=0.9\textwidth]{figure/tuanshi-1.pdf}
        \end{subfigure}
    \end{minipage}
    }
    \caption{City distribution and category distribution of 1,354,185 merchant data across 9 major cities in China.}
    \label{fig:city_and_categories}
\end{figure}

\begin{figure}[htbp] 
    \centering
    \setlength{\fboxsep}{0.1pt}
    \setlength{\fboxrule}{0.25pt}
    \resizebox{0.48\textwidth}{!}{%
    \begin{subfigure}[b]{0.33\columnwidth}
        \centering
        \fbox{\includegraphics[width=\dimexpr\textwidth-2\fboxrule\relax, height=\dimexpr0.8\columnwidth-2\fboxrule\relax]{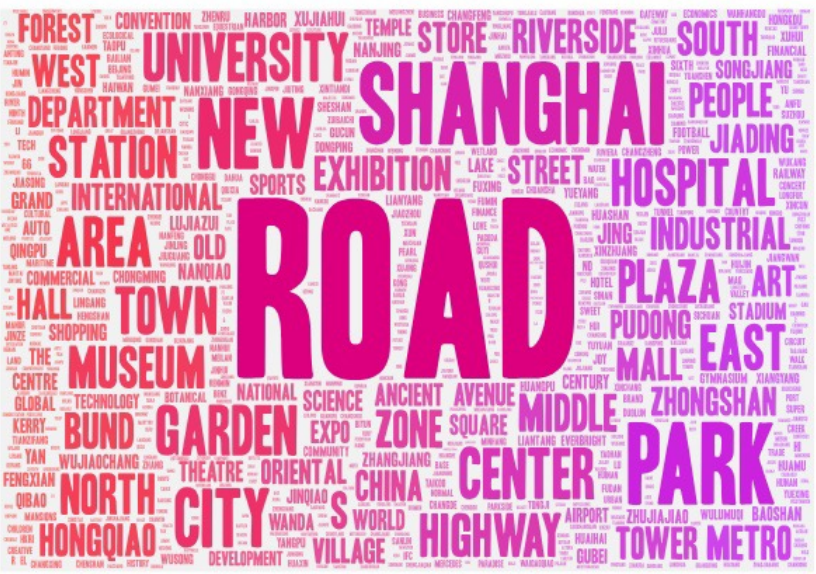}}
        \caption{Shanghai}
        \label{fig:shanghai_ciyun}
    \end{subfigure}\hspace{1mm}
    \begin{subfigure}[b]{0.33\columnwidth}
        \centering
        \fbox{\includegraphics[width=\dimexpr\textwidth-2\fboxrule\relax, height=\dimexpr0.8\columnwidth-2\fboxrule\relax]{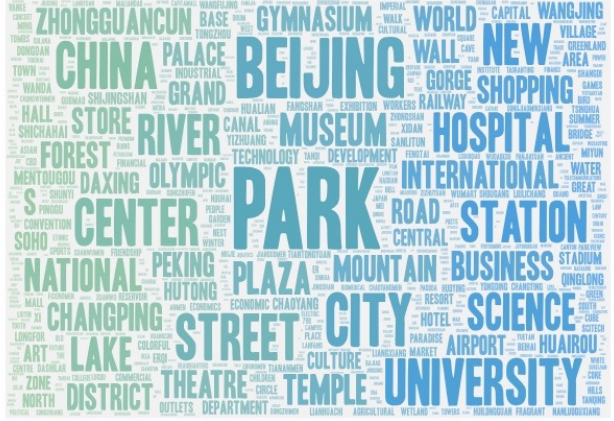}}
        \caption{Beijing}
        \label{fig:beijing_ciyun}
    \end{subfigure}\hspace{1mm}
    \begin{subfigure}[b]{0.33\columnwidth}
        \centering
        \fbox{\includegraphics[width=\dimexpr\textwidth-2\fboxrule\relax, height=\dimexpr0.8\columnwidth-2\fboxrule\relax]{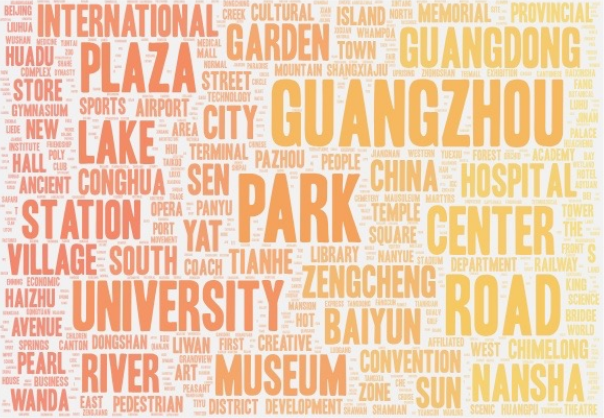}}
        \caption{Guangzhou}
        \label{fig:guangzhou_ciyun}
    \end{subfigure}%
    }
    \caption{Word clouds showing the landmark locations across 3 major Chinese cities.}
    \label{fig:city_ciyun}
    \vspace{-10pt}
\end{figure}

\subsection{Prompts for Data Construction and Validation}
\label{sec:prompts}

\begin{figure}[H]
    \centering
    \includegraphics[width=\linewidth]{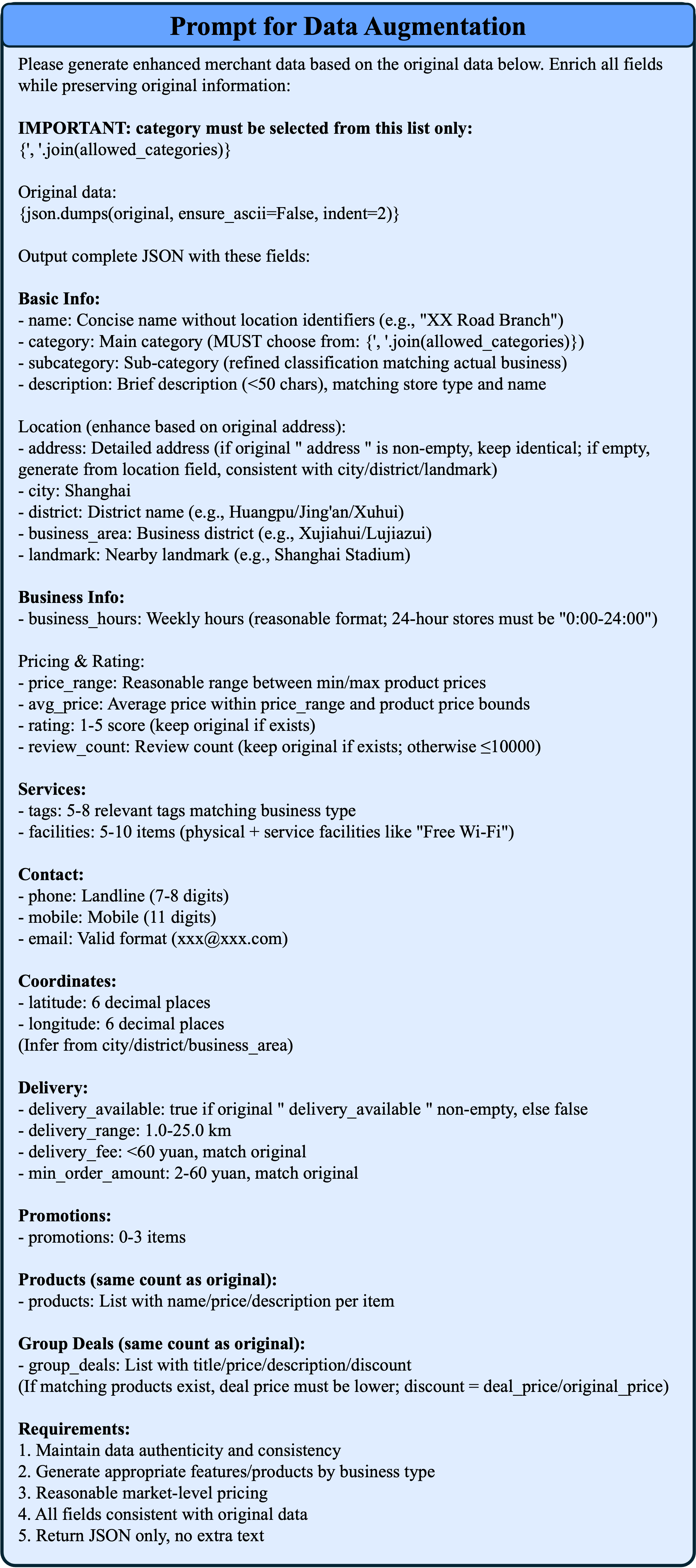}
    \caption{Prompt for Data Augmentation}
    \label{fig:prompt1}
\end{figure}

\begin{figure}[H]
    \centering
    \includegraphics[width=\linewidth]{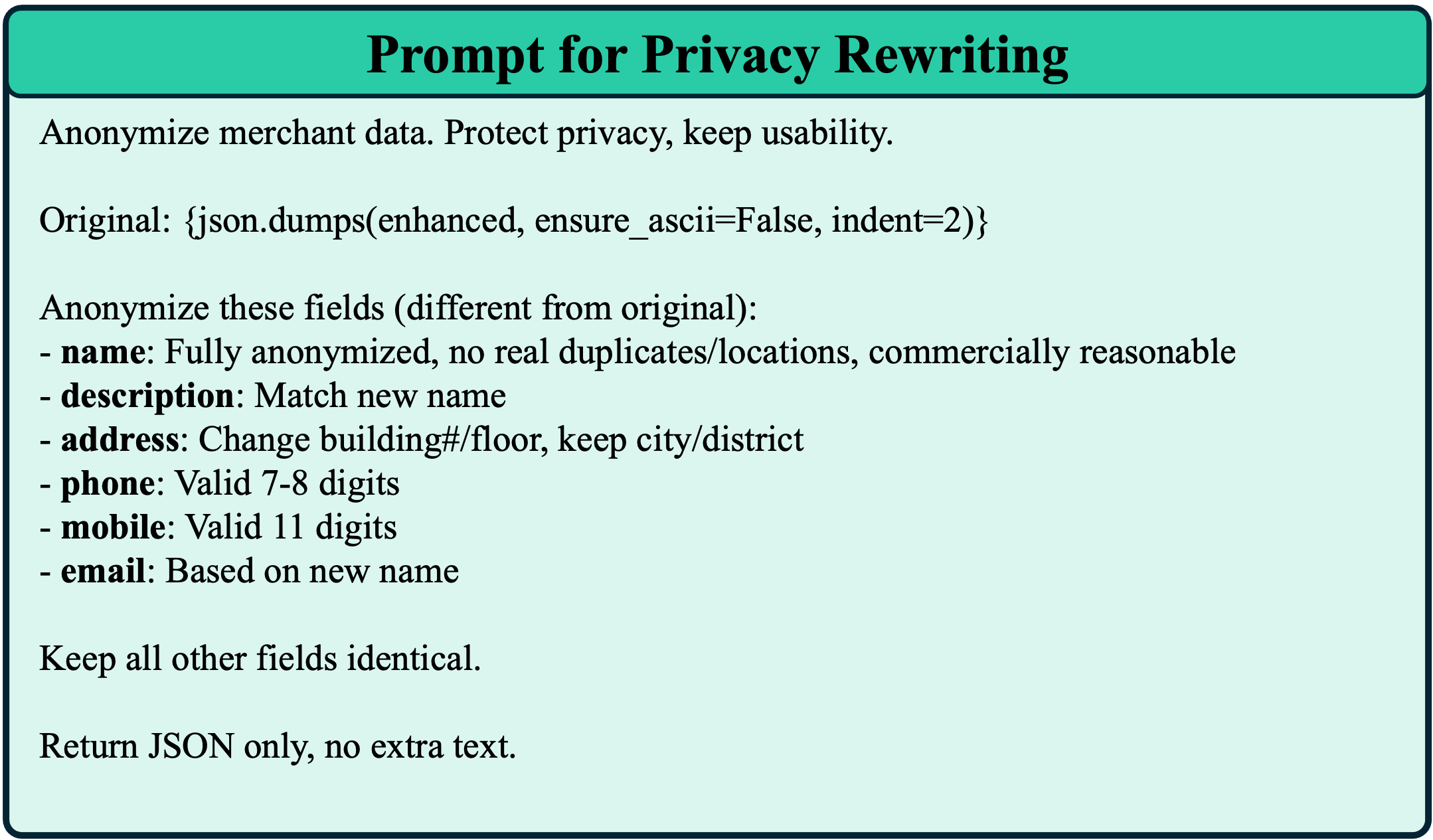}
    \caption{Prompt for Privacy Rewriting}
    \label{fig:prompt2}
\end{figure}

\begin{figure}[H]
    \centering
    \includegraphics[width=0.82\linewidth]{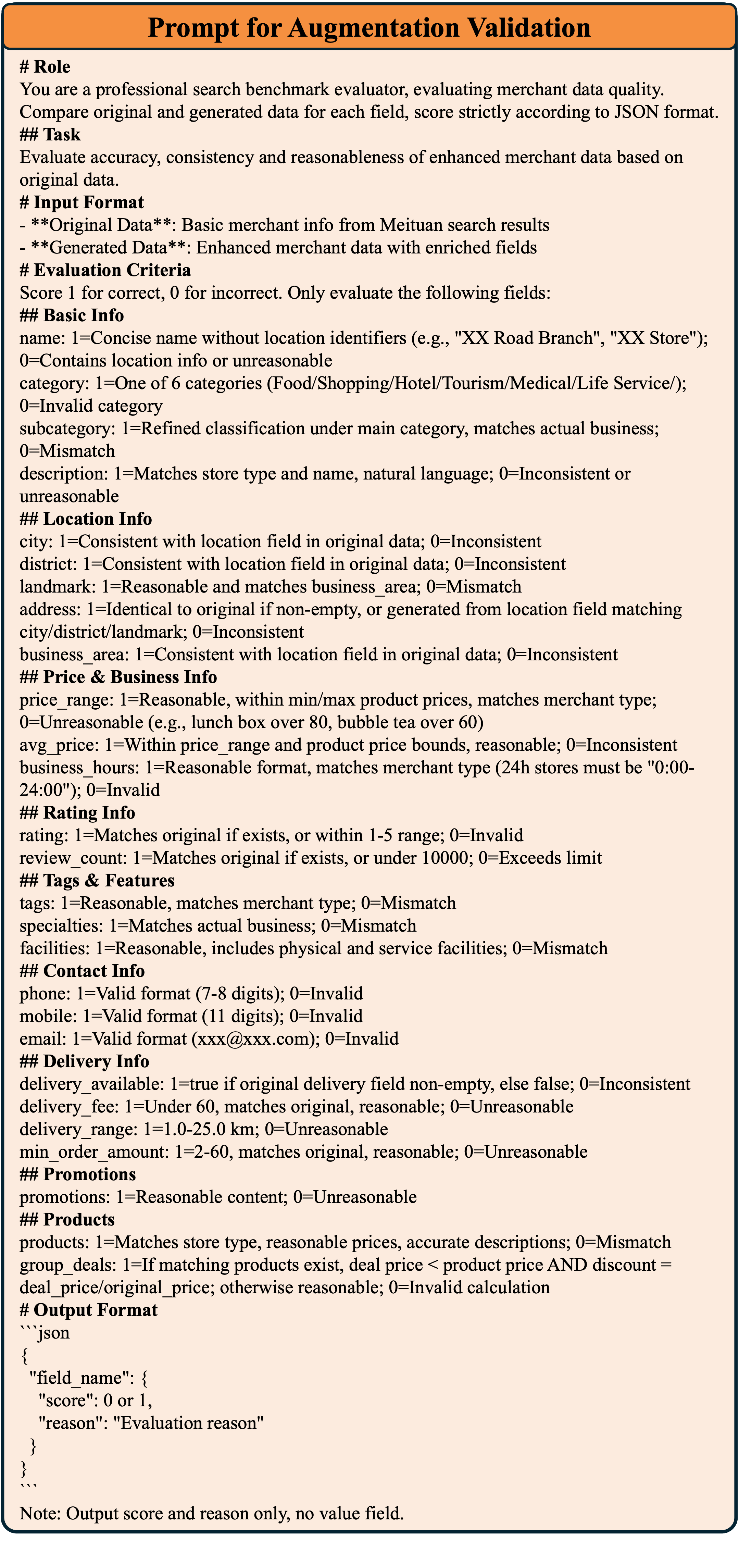}
    \caption{Prompt for Augmentation Validation}
    \label{fig:prompt3}
\end{figure}

The prompt used for data augmentation is shown in Figure~\ref{fig:prompt1}. 
The prompt for privacy rewriting is shown in Figure~\ref{fig:prompt2}. 
The validation prompt is shown in Figure~\ref{fig:prompt3}.


\end{appendices}

\end{document}